# BiomedGPT: A generalist vision–language foundation model for diverse biomedical tasks

Kai Zhang[1], Rong Zhou[1], Eashan Adhikarla[1], Zhiling Yan[1], Yixin Liu[1], Jun Yu[1], Zhengliang Liu[2], Xun Chen[3], Brian D. Davison[1], Hui Ren[4], Jing Huang[5,6], Chen Chen[7], Yuyin Zhou[8], Sunyang Fu[9], Wei Liu[10], Tianming Liu[2], Xiang Li[4*], Yong Chen[5,11,12,13], Lifang He[1*], James Zou[14,15], Quanzheng Li[4], Hongfang Liu[9], and Lichao Sun[1*]

[1]Department of Computer Science and Engineering, Lehigh University, PA, United States
[2] School of Computing, University of Georgia, GA, United States
[3]Samsung Research America, CA, United States
[4]Department of Radiology, Massachusetts General Hospital and Harvard Medical School, MA, United States
[5]Department of Biostatistics, Epidemiology, and Informatics, University of Pennsylvania, PA, United States
[6]PolicyLab, Children's Hospital of Philadelphia, PA, United States
[7]Center for Research in Computer Vision, University of Central Florida, FL, United States
[8]Department of Computer Science and Engineering, University of California, Santa Cruz, CA, United States
[9]McWilliams School of Biomedical Informatics, UTHealth Houston, TX, United States
[10]Department of Radiation Oncology, Mayo Clinic, AZ, United States
[11]The Center for Health AI and Synthesis of Evidence (CHASE), University of Pennsylvania, PA, United States
[12]Penn Institute for Biomedical Informatics (IBI), PA, United States
[13]Leonard Davis Institute of Health Economics, PA, United States
[14]Department of Biomedical Data Science, Stanford University School of Medicine, CA, United States
[15]Department of Computer Science, Stanford University, CA, United States

[*]To whom the correspondence should be addressed.
Xiang Li[4*], Email: xli60@mgh.harvard.edu;
Lifang He[1*], Email: lih319@lehigh.edu;
Lichao Sun[1*], Email: lis221@lehigh.edu.

# Abstract

Traditional biomedical artificial intelligence (AI) models, designed for specific tasks or modalities, often exhibit limited flexibility in real-world deployment and struggle to utilize holistic information. Generalist AI holds the potential to address these limitations due to its versatility in interpreting different data types and generating tailored outputs for diverse needs. However, existing biomedical generalist AI solutions are typically heavyweight and closed source to researchers, practitioners, and patients. Here, we propose BiomedGPT, the first open-source and lightweight vision-language foundation model, designed as a generalist capable of performing various biomedical tasks. BiomedGPT achieved state-of-the-art results in 16 out of 25 experiments while maintaining a computing-friendly model scale. We also conducted human evaluations to assess the capabilities of BiomedGPT in radiology visual question answering, report generation, and summarization. BiomedGPT exhibits robust prediction ability with a low error rate of 3.8% in question answering, satisfactory performance with an error rate of 8.3% in writing complex radiology reports, and competitive summarization ability with a nearly equivalent preference score to human experts. Our method demonstrates that effective training with diverse data can lead to more practical biomedical AI for improving diagnosis and workflow efficiency.

# Introduction

Artificial intelligence (AI) techniques, especially transformer-based foundation models, have demonstrated their power in solving a wide range of biomedical tasks including radiology interpretation, clinical information summarization, and precise disease diagnostics [1]. However, most of today's biomedical models act as specialist systems, tailored to specific tasks and modalities [2]. Such specialization comes with significant challenges in model deployment, especially with the growing interest in using AI for precision medicine and patient-centered care, which requires the integration and analysis of diverse data types and patient-specific details [3, 4]. Furthermore, AI's hyper-specialization in a narrow discipline often fails to provide the comprehensive insights necessary to assist human doctors in real-world settings, where the flow of information may be slow and sporadic [2, 5]. A generalist biomedical AI offers the potential to overcome these limitations by creating models that are versatile across different tasks and robust enough to handle the intricacies of varied medical data effectively [2, 6].

The emergence of general-purpose foundation models [7, 8] offers a prospective prototype for the development of biomedical generalist AI. These advanced models serialize diverse datasets, regardless of their modalities, tasks or domains, into a uniform sequence of tokens, which are then processed using a transformer neural network [9]. Unlike large language models [10, 11], which are primarily designed for processing textual data, generalist models can handle both textual and visual information simultaneously. This capability is pivotal for complex biomedical applications, where the integration of diverse data types—such as clinical text and radiographic imaging—is crucial for accurate analysis and decision-making. Furthermore, generalist models exhibit impressive multitasking capabilities, significantly simplifying the deployment and management of AI systems by reducing the need to maintain numerous narrowly-focused specialist models.

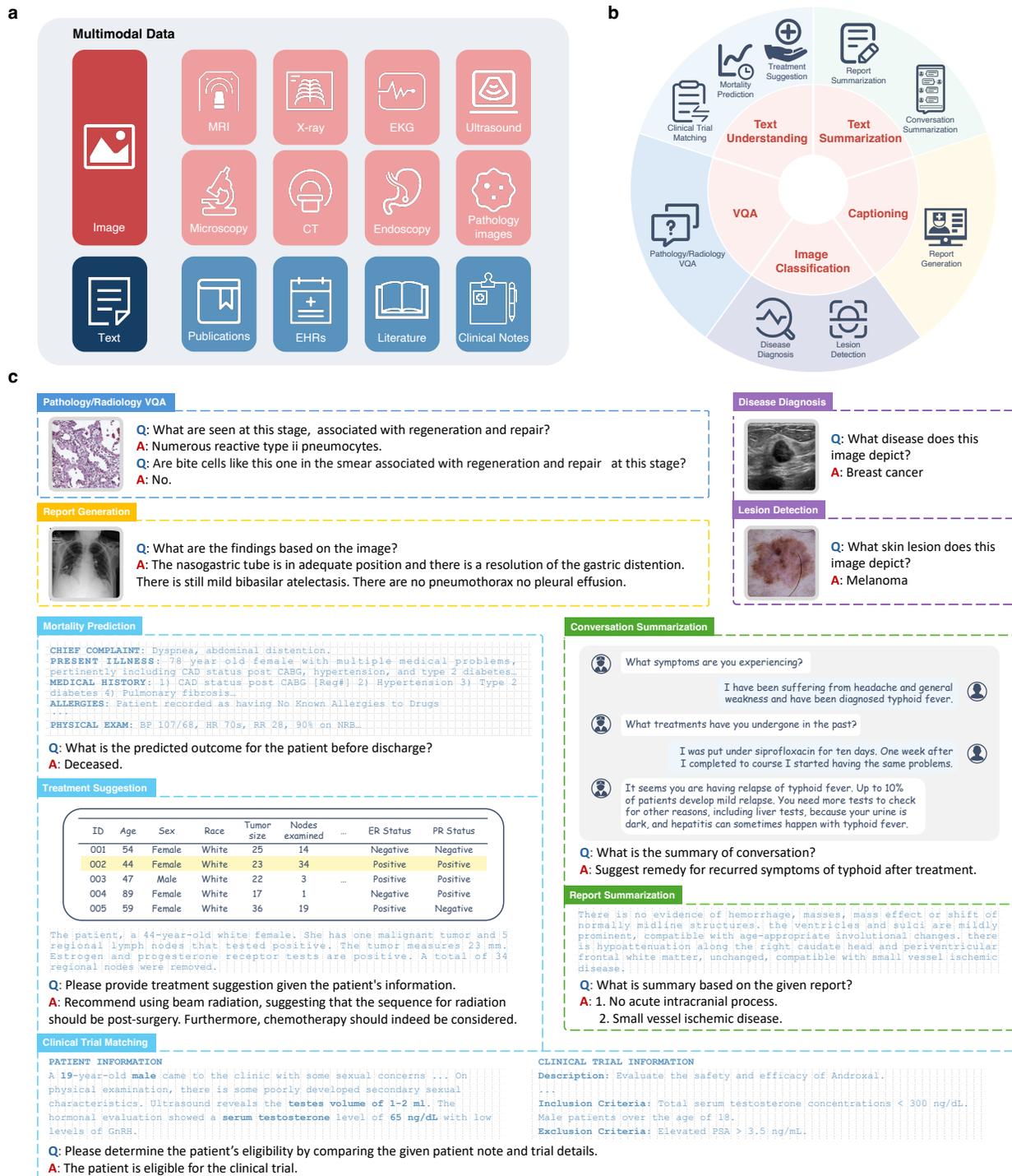

**Fig. 1**: BiomedGPT can process diverse modalities and perform versatile tasks. **(a)** BiomedGPT primarily focuses on visual and textual inputs but can also process tabular data through serialization. **(b)** Examples of the supported downstream visual-language tasks of BiomedGPT demonstrate its versatility. Additional tasks can be incorporated to meet further clinical needs via lightweight, task-specific fine-tuning. **(c)** Examples of clinical-relevant use cases with BiomedGPT include tasks, where the input may consist of both image and text or text-only, and the model responds to queries ("Q") by generating responses ("A"). Thanks to its unified framework design and comprehensive pre-training on biomedical data, BiomedGPT is highly adaptable and can be applied to a variety of downstream tasks.

In this paper, we introduce BiomedGPT, a prototype for a generalist vision-language foundation model designed to perform diverse biomedical tasks across modalities using natural language instructions (**Fig. 1**). Unlike other multimodal biomedical AI specialized for a single task [12], focused solely on one discipline [13], or not publicly accessible [6], BiomedGPT is trained with cross-disciplinary data and evaluated on a wide range of tasks. It is worth noting that our BiomedGPT is fully transparent, open-sourced, and lightweight (e.g., **3088 times smaller** than the commercial generalist biomedical AI model Med-PaLM M, which has 562 billion parameters [6]), thereby facilitating broader implementation. To empower the generalist capabilities of BiomedGPT, we curated a large-scale pre-training corpus comprising 592,567 images, approximately 183 million text sentences, 46,408 object-label pairs, and 271,804 image-text pairs (**Figs. 2c-d**). Furthermore, to enhance its instruction-following abilities, we developed a variant called Instruct-BiomedGPT with specifically curated *instruction-tuning data* (**Supplementary Fig. 1**).

To our knowledge, BiomedGPT is the first fully transparent generalist medical AI model, that is comprehensively evaluated on publicly accessible datasets and by medical professionals. This study first highlights the transfer learning capabilities of BiomedGPT, demonstrating how the model utilizes knowledge from pre-training to specialize effectively across 25 datasets through fine-tuning (**Extended Data Tables 1-2 and Supplementary Table 7**). We employed recognized metrics from previous literature to benchmark our model against state-of-the-art results. Additionally, BiomedGPT functions as a zero-shot learner, capable of answering multimodal medical questions without further training for adaptation, with performance comparable to leading AIs. Furthermore, human doctors evaluated BiomedGPT in tasks, such as VQA, report generation, and summarization within the radiology domain, demonstrating satisfactory performance. While our results highlight BiomedGPT's potential in medical applications, they also indicate that significant enhancements are required to make these models viable for real-world clinical use. Critical evaluations are particularly needed in the areas of safety, equity, and bias. Our findings underscore the challenges that must be addressed before these models can be effectively deployed in clinical settings. We outline these limitations and suggest directions for future research in this article.

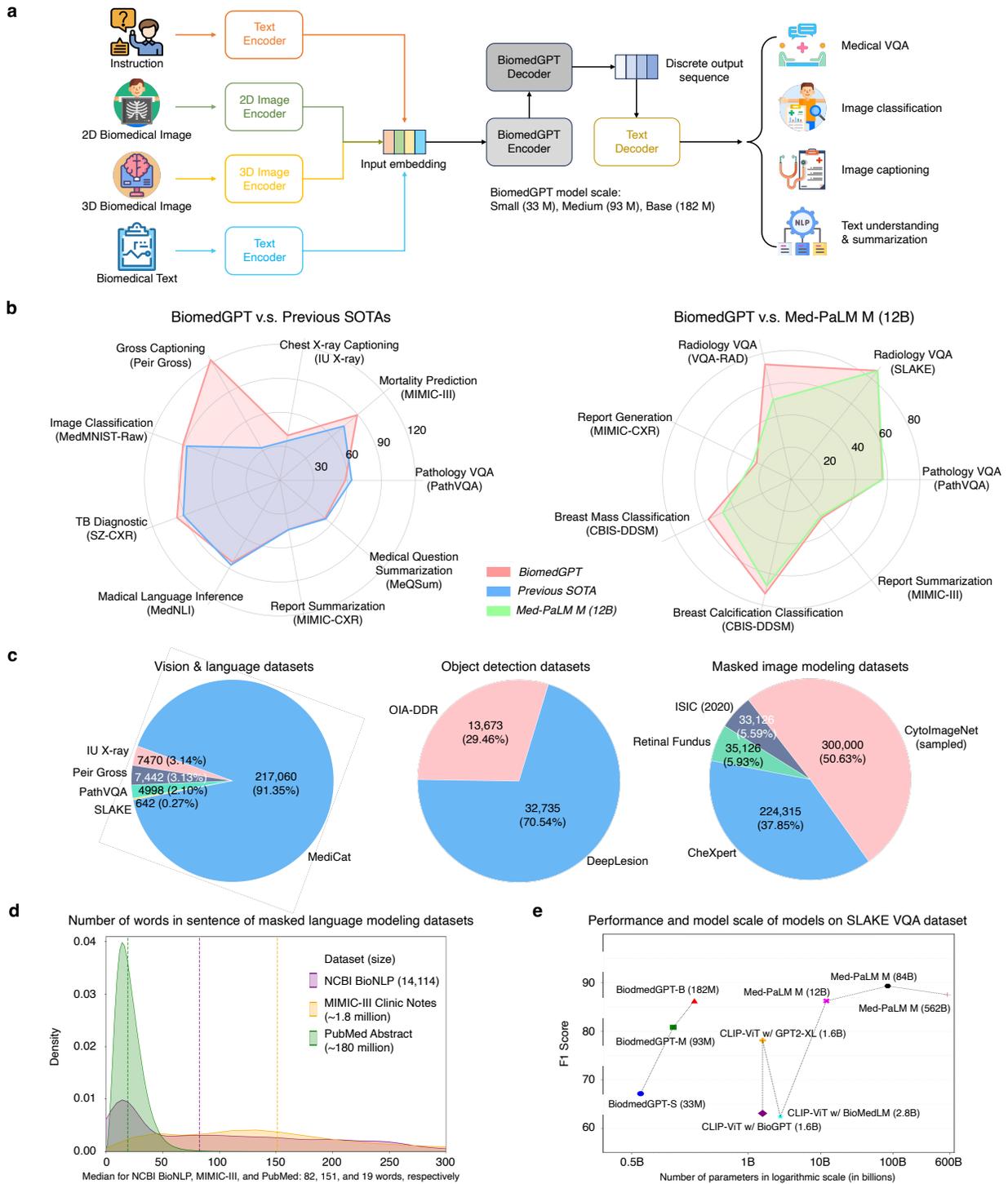

**Fig. 2**: The overview of BiomedGPT: workflow, performance, and pre-training datasets. **(a)** Graphical illustration of how BiomedGPT handles multimodal inputs and performs diverse downstream tasks. The expected form of output for each task is determined by feeding the specific instruction to the model. **(b)** Comparative performance analysis: this figure contrasts the achievements of BiomedGPT with prior SOTA results and Med-PaLM M (12B). The evaluation metrics include: accuracy for image classification, medical language inference, and visual question answering (VQA) (benchmarked against SOTA results); CIDEr for image captioning; ROUGE-L for text summarization; weighted F1 scores for VQA (in comparison with Med-PaLM M); and F1-macro for breast mass and calcification classification (also in comparison with Med-PaLM M). **(c)** Distribution of pre-training datasets including image captioning and VQA as vision & language datasets, object detection datasets, and image-only datasets for masked image modeling. **(d)** Density plot of the number of words per sentence in the text-only pre-training datasets. **(e)** Scale-related performance comparison: BiomedGPT exhibits superior performance on the SLAKE VQA dataset, even with significantly fewer parameters than its counterparts.

# Results

## Pre-training using large and diverse datasets

BiomedGPT leverages pre-training techniques including masked modeling and supervised learning, which aims to establish robust and general data representations by learning vast amounts of data across diverse tasks (**Extended Data Table 3**). To maximize the generalization of BiomedGPT, we sourced from 14 freely available datasets, ensuring the diversity of modalities (**Fig. 1a, Figs. 2c-d and Extended Data Fig. 1a**). In addition, to investigate how BiomedGPT performs across varying scales, we specifically introduced three versions of the model: BiomedGPT-S, BiomedGPT-M, and BiomedGPT-B, which correspond to small, medium, and base sizes, respectively (**Fig. 2a and Extended Data Figs. 2-3**).

## Fine-tuning for downstream tasks

Multitasking is fundamental to a generalist AI. Following previous biomedical research [14, 15, 16] and aiming for sufficiently effective performance, we mainly chose to fine-tune our model to adapt to various biomedical tasks (**Fig. 1b and Fig. 1c**). Our selection of downstream tasks stemmed from their potential real-world applications: medical image classification often aids in disease diagnostics and lesion recognition; text understanding and summarization can streamline clinic operations such as easing doctors' note-writing burdens. Furthermore, image captioning and VQA lay the groundwork for future healthcare chatbots, addressing challenges where common language might be ambiguous, but medical terminologies are too hard for most people to understand. The complete statistics of downstream datasets used in this article are shown in **Extended Data Fig. 1b**.

## BiomedGPT is lightweight but competitive in multimodal tasks

We fine-tuned BiomedGPT on two primary multimodal tasks, VQA and image captioning, each using three downstream datasets. The VQA datasets included radiology data covering five different anatomies (VQA-RAD [17] and SLAKE [18]), in addition to pathology data that captures both bodily and tissue-specific details (PathVQA [19]). For captioning, we incorporated chest X-ray datasets (IU X-ray [20] and MIMIC-CXR [21]) as well as clinical photographs from PEIR GROSS [22]. For comparison, we benchmarked BiomedGPT against leading models for each dataset [15, 23, 24, 25].

We evaluated our model's VQA performance by comparing generated answers with the ground truths. The overall accuracy of our BiomedGPT model is detailed in **Extended Data Table 1**. Notably, BiomedGPT achieved an 86.1% overall accuracy on SLAKE dataset, surpassing the previous SOTA performance of 85.4% set by BiomedCLIP [15]. Additionally, we dissected the accuracy of both "closed-ended" and "open-ended" question-answer pairs (**Fig. 3a**). Our model recorded promising closed-ended accuracies: 88.0% on PathVQA, up by 1.0% compared to the existing state-of-the-art (SOTA) model [25]. On the SLAKE dataset, BiomedGPT-B achieved an 89.9% closed-ended accuracy, down by 1.1% compared to the M2I2 model [23]. In open-ended scenarios, our model excelled with an 84.3% accuracy, surpassing M2I2's 74.7%. However, for the VQA-RAD and PathVQA datasets, BiomedGPT's performance on open-ended queries was less competitive, recording accuracies of 60.9% and 28.0%, respectively.

In addition, we compared BiomedGPT-B with Med-PaLM M (12B), using a weighted F1 score as reported in the paper. Other metrics could not be calculated due to the closed-source nature of Med-PaLM M. Remarkably, despite its much smaller size, BiomedGPT-B achieved impressive results (**Fig. 2b**). On the VQA- RAD and SLAKE datasets, BiomedGPT-B recorded scores of 73.2% and 85.2%, respectively, indicating a significant increase of 22.5% on VQA-RAD and a slight improvement of 0.02% on SLAKE. Additionally, on the PathVQA dataset, BiomedGPT-B scored a weighted F1 of 56.9%, only marginally lower by 0.4% than Med-PaLM M, while utilizing a model with 98.5% fewer parameters.

For image captioning (**Fig. 3b**), we meticulously assessed the quality of machine-generated text using three metrics: ROUGE-L [26], METEOR [27], and CIDEr [28], following baseline methods [13, 29, 30, 31, 32, 33]. These evaluation metrics are useful for assessing the similarity and consensus between the generated text and the reference text written by medical experts. They have also shown some alignment with physician raters [34]. Consequently, models that score higher on these natural language processing (NLP) metrics can be selected as candidates for further human evaluations [35]. On the PEIR GROSS dataset, our BiomedGPT model surpassed the existing SOTA benchmark [36], demonstrating improvements of 8.1 percentage points in ROUGE-L, 0.5 points in METEOR, and a significant gain of 89.8 points in the CIDEr metric. Conversely, on the IU X-ray dataset, BiomedGPT achieved a leading CIDEr score of 40.1, marking a 5.0-point improvement over the SOTA model [31]. On the MIMIC-CXR dataset, in terms of METEOR, our model recorded a score of 15.9%, surpassing the previous leading result set by [30].

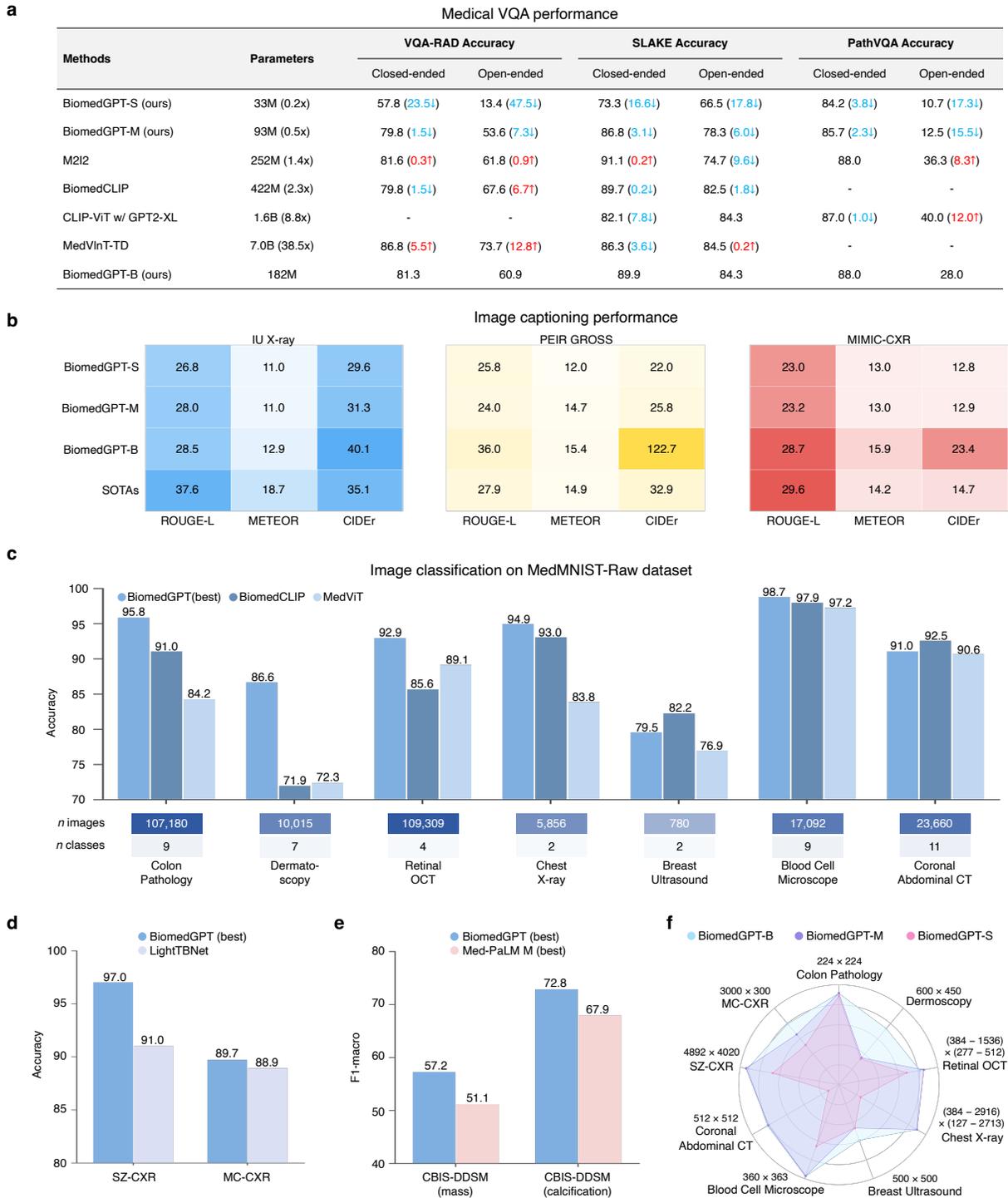

**Fig. 3**: BiomedGPT performs fine-tuning for vision-language and medical image classification downstream tasks. **(a)** Medical VQA performance of BiomedGPT and the leading models in terms of closed-ended and open-ended accuracies. **(b)** Image captioning performance of BiomedGPT and SOTAs on IU X-ray, PEIR GROSS and MIMIC-CXR data. The evaluation metrics are ROUGEL-L, METEOR and CIDEr. **(c)** Evaluation of image classification on the MedMNIST-Raw dataset for each domain type. **(d)** Image classification performance with accuracy across two super-resolution image datasets. **(e)** Image classification performance with F1-macro on CBIS-DDSM dataset. **(f)** Accuracies across 9 datasets with different resolutions vary in terms of model scales. In general, larger models tend to perform better.

## BiomedGPT enables accurate medical image classification

For medical image classification task, we curated seven biomedical image datasets encompassing seven modalities following [37], denoted as MedMNIST-Raw: (1) colon pathology with nine tissue types; (2) dermatoscopy images of seven typical pigmented skin lesions; (3) breast ultrasound (normal, benign, and malignant); (4) retinal OCT categorized into four types of retinal diseases; (5) chest X-ray images for binary-class classification of pneumonia against normal; (6) blood cell microscope showcasing eight kinds of normal cells; (7) Abdominal CT with 11 body organs across coronal view. Additionally, we tested the model on two super-resolution pulmonary disease datasets, with a specific focus on pulmonary tuberculosis (TB), which has a limited number of samples: (8) Montgomery County chest X-ray set (MC-CXR) with the size of either 4020×4892 or 4892×4020 pixels; (9) Shenzhen chest X-ray set (SZ- CXR) with approximate dimensions of 3000×3000 pixels. To be consistent with prior works, we employed accuracy for evaluation. As shown in **Figs. 3c-e**, BiomedGPT outperformed previous SOTAs on 7 of the 9 biomedical image classification datasets after 5-epoch fine-tuning.

Notably, on the SZ-CXR and MC-CXR datasets [38] (binary classification), BiomedGPT posted accuracies of 97.0% and 89.7%, reflecting improvements of 6.0% and 0.8% over the previously leading model – LightTBNet [39], respectively (**Fig. 3d**). For MedMNIST-Raw, we selected two top-performing approaches on biomedical imaging analysis, MedViT (Large) [40] and BiomedCLIP [15], as benchmarks to compare with BiomedGPT. For BiomedCLIP, we added a decision layer and fine-tuned the entire model. BiomedGPT achieved 5 out of 7 best accuracies on MedMNIST-Raw (**Fig. 3c**), for example, on the dermatoscopy dataset, BiomedGPT surpassed the two baseline models by more than 14%. On average, BiomedGPT achieved performance improvements of 6.1% and 3.3% over MedViT and BiomedCLIP, respectively.

Considering the model scale, it becomes evident that BiomedGPT exhibits performance enhancements as its scale increases (**Fig. 3f**). Specifically, on the MC-CXR dataset, the small model logged 75.9% accuracy. In contrast, the medium model showcased a score of 82.8%, which is 6.9% higher than its smaller counterpart, and the base model continued this upward trajectory with a score of 89.7% surpassing the medium model by 6.9%. However, we also observed performance saturation on several datasets such as SZ-CXR. We also tested the extreme situation where the images were resized to a very small scale and found that performance saturation became much more pronounced (**Supplementary Table 1**).

Additionally, we benchmarked our BiomedGPT against Med-PaLM M on the CBIS-DDSM dataset [41] for both 3-class lesion-level mass classification and calcification classification. Using the F1-macro as the evaluation metric, consistent with the Med-PaLM M, we found that BiomedGPT-B outperforms all versions of Med-PaLM M, spanning 12B, 84B, and 584B parameters (**Fig. 3e and Extended Data Fig. 4a**). These findings underscore the impressive efficiency and efficacy of BiomedGPT, even when squared against models of larger scales.

## BiomedGPT is able to understand and summarize clinical text

We assess the BiomedGPT's proficiency in understanding and condensing complex medical narratives that hold potential for real-world clinical needs: (1) medical natural language inference, utilizing the MedNLI dataset [42], which tests the model's comprehension in deducing hypotheses from provided premises; (2) treatment suggestions for radiation therapy and chemotherapy based on Surveillance, Epidemiology, and End Results (SEER) dataset [43]. (3) in-hospital mortality prediction based on admission notes; (4) clinical-trial matching that identifies lists of candidate clinical trials suitable for individual patients. Moreover, we explored the BiomedGPT's performance in medical text summarization, which was applied to datasets of doctor-patient dialogues (MedQSum [44] and HealthCareMagic [45]) as well as radiology reports (MIMIC-CXR [21] and MIMIC-III [46]).

In the evaluation of the MedNLI dataset for 3-class classification (entailment, contradiction, or neutral), we adopted accuracy as our evaluation metric, consistent with prior research (**Fig. 4e**). Notably, when compared to the state-of-the-art performance of SciFive-Large [16] at 86.6% accuracy, our BiomedGPT-B, with merely a quarter of SciFive-Large's parameter count, exhibited a modest decline in accuracy of only 2.8%.

For the treatment suggestion task, we adopted the preprocessing steps from the data-sourced work [47]. An example output is like: "*Recommend using beam radiation, suggesting that the sequence for radiation should be post-surgery. Furthermore, chemotherapy should indeed be considered.*" To evaluate the effectiveness of three variants in treatment suggestions, we employed a 10-fold cross-validation method and compared its current open-source state-of-the-art methods, including BioGPT [14] and LLaVA-Med (using the language backbone) [12] (**Fig. 4a**), which have 347M and 7B parameters, respectively —approximately 11 and 212 times larger than BiomedGPT-S. BiomedGPT-B achieved a mean accuracy of 50.0% ± 5.3%, outperforming BioGPT and LLaVA-Med, which had accuracies of 45.9% ± 4.8% and 41.5% ± 7.1%, respectively. Considering the complexity involved with six types of radiation therapy, seven radiation sequences and two types of chemotherapy [49], which together imply a random guess accuracy of 1.2%, both BiomedGPTs and the baseline models demonstrate significantly higher accuracies than this baseline.

To assess BiomedGPT's performance in predicting in-hospital mortality, we used admission notes extracted from the MIMIC-III database, following [48], with the official test set. **Fig. 4c** presents the prediction accuracy results for five models, demonstrating that all three versions of BiomedGPT outperform BioGPT and LLaVA-Med. Notably, BiomedGPT-B achieves an accuracy improvement of over 15% compared to these two baselines.

For the clinical-trial matching task, we collected a dataset from TREC 2022 [49], categorized into three groups: eligible, irrelevant, and ineligible. We randomly chose 80% from each group as a training set and the remaining 20% as a test set and reported the average results across 10 repetitions. Again, all three versions of BiomedGPT outperform the baselines (**Fig. 4b**). In particular, BiomedGPT-B achieved a mean accuracy of 85.2% ± 1.5%, significantly outperforming BioGPT and LLaVA-Med, which only had accuracies of 42.0 % ± 1.8% and 48.7% ± 2.4%, respectively.

In evaluating text summarization, we employed the ROUGE-L metric to assess BiomedGPT-B's performance across four benchmark datasets (**Fig. 4d**). BiomedGPT-B demonstrated its capability in summarizing doctor-patient dialogues on the MedQSum and HealthCareMagic datasets, achieving ROUGE-L scores of 52.3% and 42%, respectively. When compared to leading models [32] with 400 million parameters (2+ times larger than BiomedGPT-B) that recorded ROUGE-L scores of 53.2% and 44.7%, BiomedGPT-B showed only minor performance drops of 0.9% and 2.7%. Additionally, in summarizing radiology reports, specifically in generating impressions from radiologists' findings, BiomedGPT-B achieved a ROUGE-L score of 44.4% on the MIMIC-CXR dataset. This result is closely aligned with the state-of-the-art, trailing by a mere 0.1% from the top score of 44.5% [50]. In the MIMIC-III dataset, BiomedGPT-B's performance stood out with a ROUGE-L score of 30.7%, surpassing Med-PaLM M (12B) which scored 29.5%.

## BiomedGPT can perform zero-shot prediction on new data

In our study, we focused on evaluating the zero-shot capabilities of BiomedGPT in VQA, highlighting its ability to answer biomedical questions in a freeform manner at scale, without requiring retraining. This contrasts sharply with earlier biomedical AI models, such as BERT-based or ViT-based models [40] that are incapable of zero-shot prediction, or CLIP-based models [15], which necessitate predefined answer (**Extended Data Fig. 5a**). Unlike these models, BiomedGPT can generate answers by simply processing the input data, offering more flexible and dynamic AI-driven solutions for biomedical inquiries. In addition to medical VQA, BiomedGPT showcased zero-shot capabilities in disease diagnosis and X-ray report generation, matching the performance of Med-PaLM M and LLaVA-Med (**Extended Data Figs. 5b-c**).

We conducted our evaluation on VQA-RAD dataset [18] (absent from the pre-training data) through 50 random samplings. Our performance evaluation of BiomedGPT centered on two key metrics: (1) the accuracy of the model in providing correct answers, and (2) its ability to understand the questions and respond in a contextually relevant manner, measured as alignment accuracy. We noted low alignment accuracy indicating poor question comprehension by our pre-trained models (**Fig. 4f**). To address this, we developed Instruct-BiomedGPT by fine-tuning with instruction-tuning data (**Supplementary Fig. 1**). We assessed this model against current SOTA models including GPT-4V [51], LLaVA-Med (7B) [12], OFA-Huge (930M), and OFA-Large (470M) [52] in a zero-shot setting, analyzing various question types (**Extended Data Table 4**). Specifically, Instruct-BiomedGPT-B achieved a zero-shot accuracy of 54.7% ± 5.7% surpassing GPT-4V's 53.0% ± 6.7% (**Fig. 4h**). Despite this improvement in understanding medical questions, neither model reached clinically acceptable performance. For example, the currently top-performing medical vision-language model, LLaVA-Med, achieved only 42.0% and 40.6% accuracies in disease diagnosis and lesion detection, respectively (**Fig. 4g**). While Instruct-BiomedGPT-B showed a substantial improvement over LLaVA-Med by more than 10%, accuracies remained under 60%. These results highlight the complexity of diagnosis and the need for ongoing fine-tuning in the development of visual-language biomedical AI.

Regarding alignment accuracy, GPT-4V and LLaVA-Med outperform other models (**Fig. 4f**). Specifically, they achieved an impressive score of 99.5% ± 1.1% and 98.2% ± 2.0% respectively, likely due to the advanced large language models upon which they are built [10, 11]. The marked improvement in alignment accuracy between Instruct-BiomedGPT and the pre-trained BiomedGPT exemplifies the effectiveness of instruction tuning in enhancing the model's capability to follow instructions accurately. For instance, while BiomedGPT-B achieved a mean alignment accuracy of 79.2%, Instruct-BiomedGPT-B reached 95%.

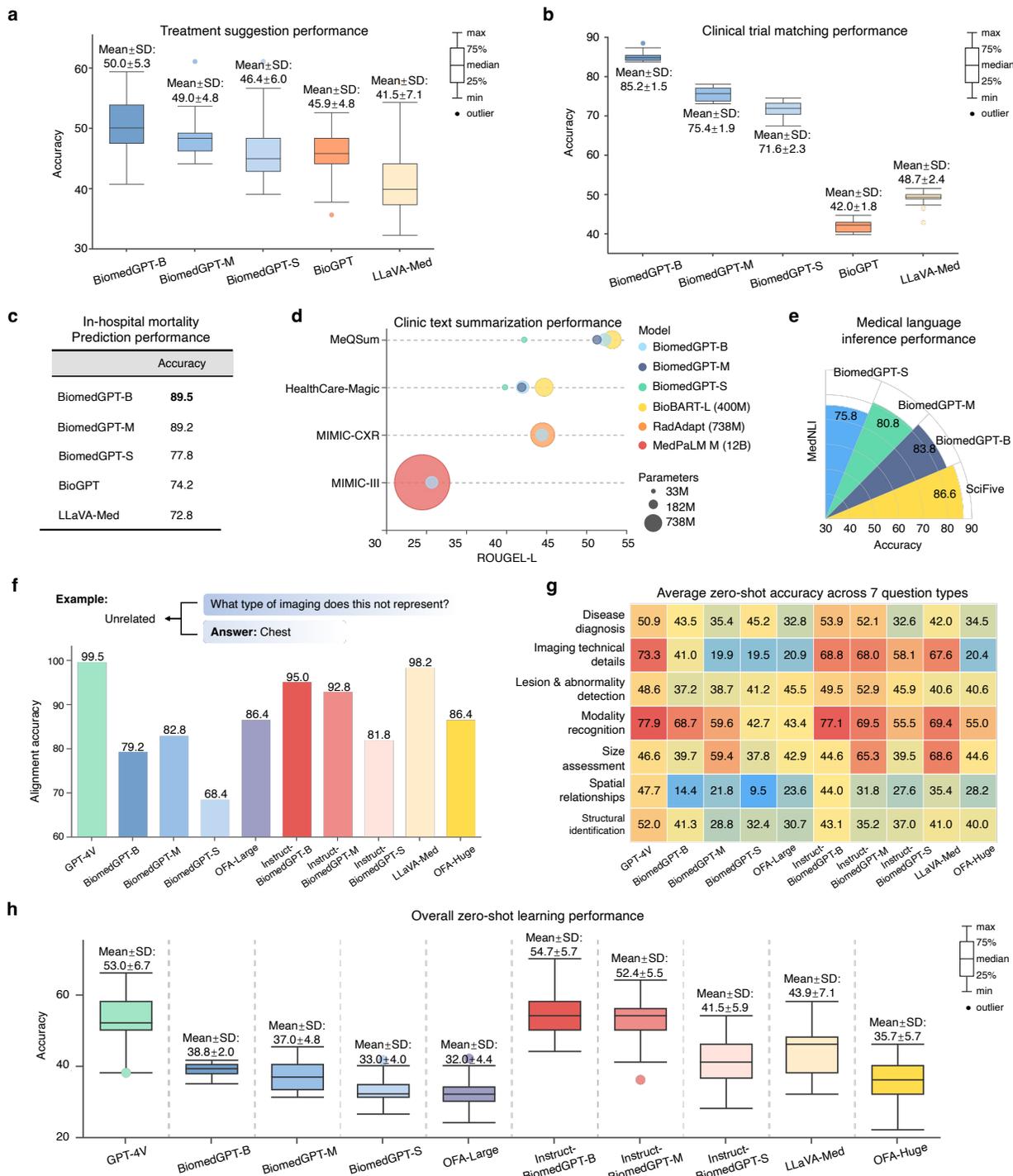

**Fig. 4**: BiomedGPT performs few-epoch transfer learning for clinic text understanding and summarization and generates the response via zero-shot transfer learning. **(a)** Evaluation of models for treatment suggestion task in terms

of accuracy using 10-fold cross-validation (*n*=4680). **(b)** Performance comparison with accuracy on the patient-trial matching dataset derived from TREC 2022 dataset using 10-fold cross-validation (*n*=7079). **(c)** Accuracy across three BiomedGPT variants and two SOTA models, BioGPT and LLaVA-Med for in-hospital mortality prediction. **(d)** ROUGEL-L across four text summarization datasets in terms of model scales. The size of the legends represents the number of parameters. **(e)** Medical language inference performance on MedNLI dataset. **(f)** Zero-shot question alignment accuracy comparison among Instruct-BiomedGPTs (base, medium, small), BiomedGPTs, OFAs (large, huge), LLaVA-Med and GPT-4V. We provide an example illustrating a mismatch between the generated answer and the question. **(g)** Average zero-shot accuracy across 7 question types on VQA-RAD dataset. **(h)** Overall zero-shot learning performance on VQA-RAD dataset through 50 repeated sampling (*n*=39).

## Human evaluation of BiomedGPT for radiology tasks

To evaluate the clinical applicability and deployment challenges of BiomedGPT, we conducted a series of analyses through radiologist evaluations of the model's generated responses to a wide range of tasks including VQA, report generation and report summarization in radiology. Examples of human evaluation on these three tasks in terms of response factuality, omissions, and significance of the errors are shown in **Fig. 5a**. The detailed evaluation procedure and performance analysis are described as follows:

**Radiology VQA.** To clinically evaluate the correctness of BiomedGPT's responses, we randomly selected 52 question-answer samples from 16 images in the official test set of MIMIC-Diff-VQA [53] over six categories (**Supplementary Table 2**): abnormality, presence, location, type, view, and severity level. For a fair comparison, we collected the answers generated by BiomedGPT, LLaVA-Med after fine-tuning and GPT-4V (zero-shot). The generated answers were presented to the seasoned radiologist at MGH for scoring (**Figs. 5b-c**). The answers were categorized as correct, partially correct, incorrect, or unrelated, and assigned them scores of 2, 1, 0, and -1, respectively. Additionally, the original radiology reports will be provided to the radiologist to serve as a reference, facilitating a potentially more precise evaluation.

BiomedGPT achieved an average score of 1.75 across all 52 samples, accumulating a total score of 91. In comparison, GPT-4V and LLaVA-Med attained an average score of 1.17 and 1.4, resulting in a total score of 61 and 73, respectively. When considering the types of questions, BiomedGPT demonstrated superior performance in four out of five categories. In addition, despite our radiologist identifying some errors in the sampled gold labels from MIMIC-Diff-VQA, we established a comparison using an exact match score based on these labels across the test set with non-difference questions. In this evaluation, BiomedGPT-B showed the best performance (**Supplementary Table 3)**.

**Radiology report generation.** This task's complexity arises from the need for long-form outputs that provide detailed descriptions of various aspects such as the presence, location, and severity of abnormalities. In this study, we randomly selected 30 sample image-report pairs from the MIMIC-CXR dataset [21]. We then applied BiomedGPT-B and BiomedGPT-M to generate the "findings" section of the radiology report based on the input chest X-ray (CXR) image. The radiologist will assess the quality of the generated text by addressing several aspects. First, they will identify any disagreements with the generated report, such as incorrect finding locations, incorrect severity levels, references to views not present, or mentions of prior studies that do not exist. Second, the radiologist will determine whether the errors in the generated report are significant, with options being significant, insignificant, or N/A if more information is needed. Third, they will pinpoint any omissions in the generated text. Finally, the radiologist will judge whether these omissions are clinically significant.

In such an evaluation, we focused on finding-level metrics, where the generated text would be split into individual findings. For instance, the report "*PA and lateral views of the chest provided. Cardiomegaly is again noted with mild pulmonary edema. No large effusion or pneumothorax.*" consists of three distinct findings. To clearly demonstrate the quality of the generated findings, we have quantified the error rates and omission rates (**Fig. 5d**). In the analysis of 192 generated findings, BiomedGPT-B achieved a rate of "significant error" of 8.3%, while BiomedGPT-M exhibited a rate of 11.0% (excluding one case that requires additional information for a comprehensive impact assessment). These rates are comparable to the human observer variabilities on the MIMIC-CXR, which has an error rate of approximately 6% [54]. We also reported the rate of "harmless error", where BiomedGPT-B and BiomedGPT-M achieved 5.2% and 11.5%, respectively. Our observations included an analysis of 254 findings from the reference report to calculate the omission rates. The total omission rate for BiomedGPT-B and BiomedGPT-M was 23.3% and 23.5%, respectively. Since not all findings described in the reference are clinically necessary, our analysis primarily focused on significant omissions, where both models had similar rates of 7.0% and 6.9%, respectively.

**Radiology report summarization.** We evaluated 100 summaries generated by BiomedGPT-B based on findings from MIMIC-CXR data [21], alongside the "Impression" sections of corresponding reference reports. Our evaluation focused on completeness, correctness, and potential medically adverse effects due to any omissions or incorrect interpretations (**Fig. 5a**). Completeness is rated from 1 (very incomplete) to 5 (very complete), with 3 representing a borderline (neutral) encapsulation. Accuracy is assessed by how well the content reflects the clinical implications for the patient, rated from 1 (very incorrect) to 5 (very correct). The potential for medically adverse effects from errors is classified as "no harm", "mild", or "severe", based on their clinical impact. Finally, we compared which summary, generated or referenced, better encapsulates all clinically relevant information, providing a comprehensive comparison of AI-generated summaries to traditional radiology reports in terms of relevance, accuracy, and safety.

BiomedGPT-generated summaries generally exhibit higher completeness (**Fig. 5e**), achieving average completeness (score > 3) in 81.0% of cases, 15.0% higher than the reference summaries. Additionally, only 5% of BiomedGPT-generated summaries are considered incomplete (score < 3), compared to 4% for the reference summaries. Despite these, the average completeness score for BiomedGPT is slightly lower at 3.9, versus 4.0 for reference summaries, with no significant difference (p > 0.05). BiomedGPT also demonstrated a higher correctness rate, with 90.0% of its summaries scoring above 3, compared to 86.0% for the reference impressions. The Wilcoxon rank-sum test showed no significant difference (p > 0.05) in average correctness scores between BiomedGPT and the reference summaries, both averaging 4.4 out of 5. In addition, our analysis found that 6.0% of BiomedGPT-generated summaries contained medically adverse items, categorized as either "mild" or "severe", which is identical to the rate observed in the reference impressions. This indicates that BiomedGPT-generated summaries are comparable to human experts in terms of medical safety. Notably, there was one instance of a "severe" adverse effect identified in the reference impressions, with no such cases found in the BiomedGPT-generated summaries. The overall score of summaries generated by BiomedGPT closely matches those produced by the reference, with preference scores of 48% for BiomedGPT and 52% for the reference (**Fig. 5e**). The results of the Sign test ($p > 0.05$) indicate no significant preference for either system, suggesting comparable performance in delivering quality and safety in medical summarization.

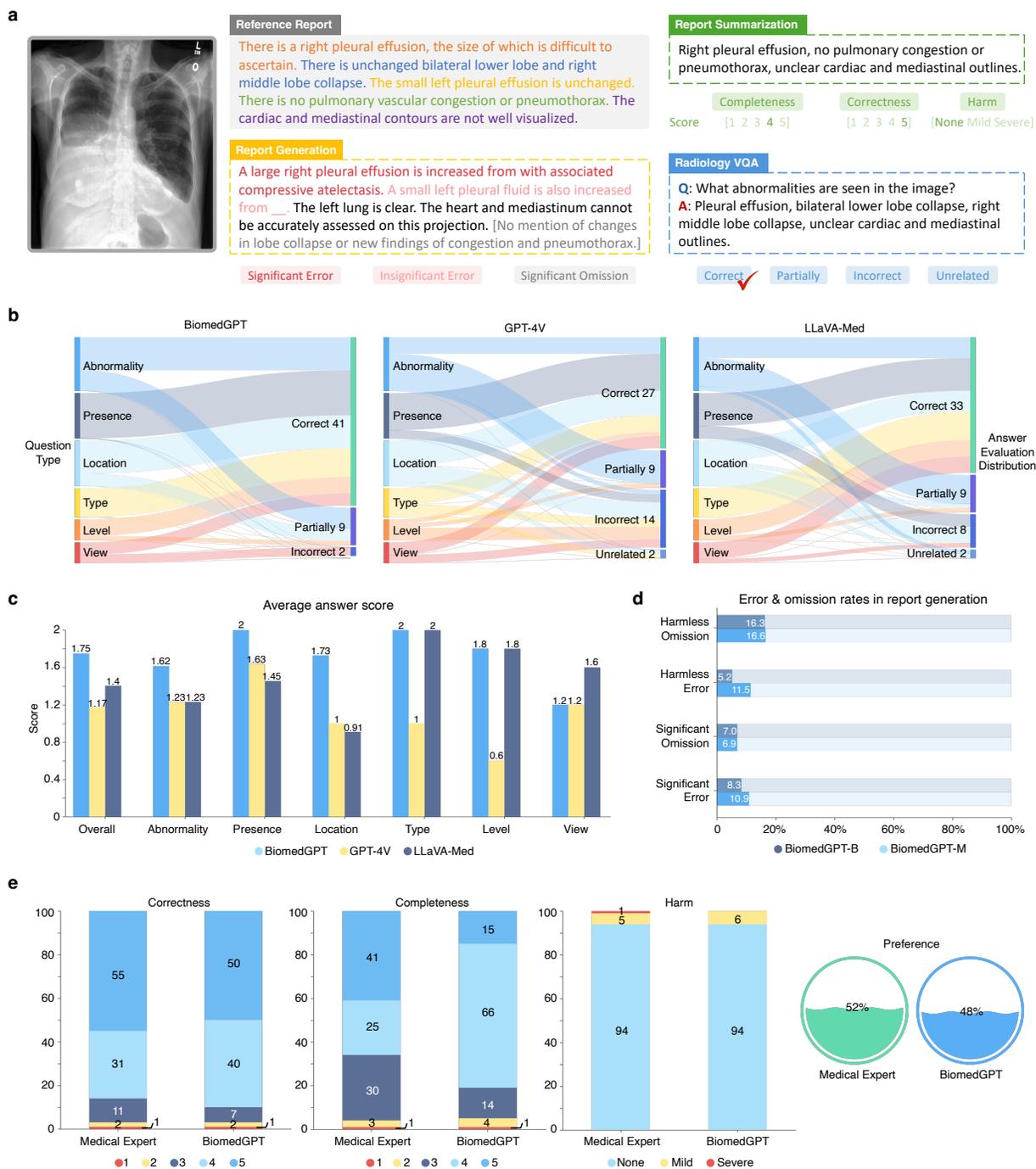

**Fig. 5**: Human evaluation of the VQA, text summarization, and captioning tasks. **(a)** Examples of human evaluation for three tasks in terms of response factuality, omissions, and significance of the errors. **(b)** Performance comparison

between three models across six question categories for radiology VQA. **(c)** Average answer score for radiology VQA. **(d)** Error and omission rates of BiomedGPT-B and BiomedGPT-M in the generated radiology report. **(e)** Human evaluation of report summarization considers three attributes: completeness, correctness, and potential harm, with an overall preference.

# Discussion

In this study, we have shown that BiomedGPT can achieve competitive transfer learning performance across vision, language, and multimodal domains by integrating diverse biomedical modalities and tasks within a unified pre-training framework. However, the experimental results also revealed limitations, offering insights for potential improvement.

The development of AI is critically dependent on the availability of high-quality and annotated data. This requirement poses a unique challenge in the biomedical domain, where data annotation is expensive, time-consuming, and demands extensive domain [55]. Consequently, AI researchers often resort to public datasets, which may compromise data quality. When dealing with multimodal biomedical datasets, particularly image-text pairs, these issues become more pronounced: (1) most existing datasets focus primarily on radiology, contributing to a significant modality imbalance; (2) the data scale of the image with detailed annotation is still limited compared to unlabeled or weakly-labeled biomedical images, and the accessible biomedical articles from PubMed or PMC. In our study, we have considered diverse modalities and ensured sufficient data scale to train high-performance models. As more biomedical data are curated and open-sourced, we can obtain better visual-semantic mappings (**Fig. 6**).

Evaluating the quality of generated text presents significant challenges. While metrics such as CIDEr and ROUGE-L can measure the agreement between generated content and the gold standard, and are commonly used for model selection to further assess their clinical applicability [35], ensuring the factual accuracy of these outputs remains a concern. To address this, recent research introduced the F1-RadGraph score [56], which qualitatively assesses the factual correctness and completeness of generated reports. In other domains like pathology, however, similar evaluation metrics are not yet prevalent. Drawing inspiration from factual-concerned metrics developed in radiology [58], we anticipate the emergence of analogous metrics for these domains. These would further enhance our ability to measure the factual integrity and overall quality of AI-generated medical content across various biomedical fields.

BiomedGPT currently adept in processing images and texts, could potentially extend its capabilities to other types of biomedical data, such as video and time-series/sequential data. For instance, we demonstrated extending BiomedGPT to handle 3D images by introducing a 3D image encoder into the framework (**Extended Data Table 5 and Supplementary Table 4)**. Nevertheless, these expansions raise concerns about negative transfer, where learning from additional modalities might inadvertently hamper performance on certain tasks. For instance, our ablation study revealed that excluding image data during pre-training improves performance on language-only downstream tasks (**Fig. 6a**), highlighting the risk of negative transfer. To mitigate this, we propose exploring controllable learning strategies such as the mixture of experts [57].

Evidence from our comprehensive analysis (**Figs. 3a-b, Fig. 3f, Figs. 4a-e, and Fig. 4h**) indicates a direct correlation between increased model scale and enhanced performance, applicable to both zero-shot predictions and post-fine-tuning. However, scaling brings its own set of challenges, particularly concerning fine-tuning efficiency, training speed, and memory requirements. We have tried to address the efficiency challenges of BiomedGPT by exploring prompt tuning, which adds small-scale parameters to condition-frozen models [58]. However, this method incurred large performance degradation (**Extended Data Fig. 4b**).

Our zero-shot transfer learning tests (**Figs. 4f-h**) indicated that BiomedGPT's text comprehension capabilities, especially in comparison with GPT-4V, are not fully established. Two main factors contribute to this limitation: Firstly, the current scale of BiomedGPT especially the language backbone is limited by available resources, though it is expandable. Our preliminary observations indicate that even with 7 billion parameters and effective training, achieving robust zero-shot in-context or text understanding remains challenging in complex medical applications. However, fine-tuning, even with a smaller-scale model like BiomedGPT, proves to be a promising approach to mitigate risks (**Supplementary Fig. 3**). Secondly, the use of a single encoder that handles multiple input types complicates the separation of diverse modality representations, requiring more refined training strategies.

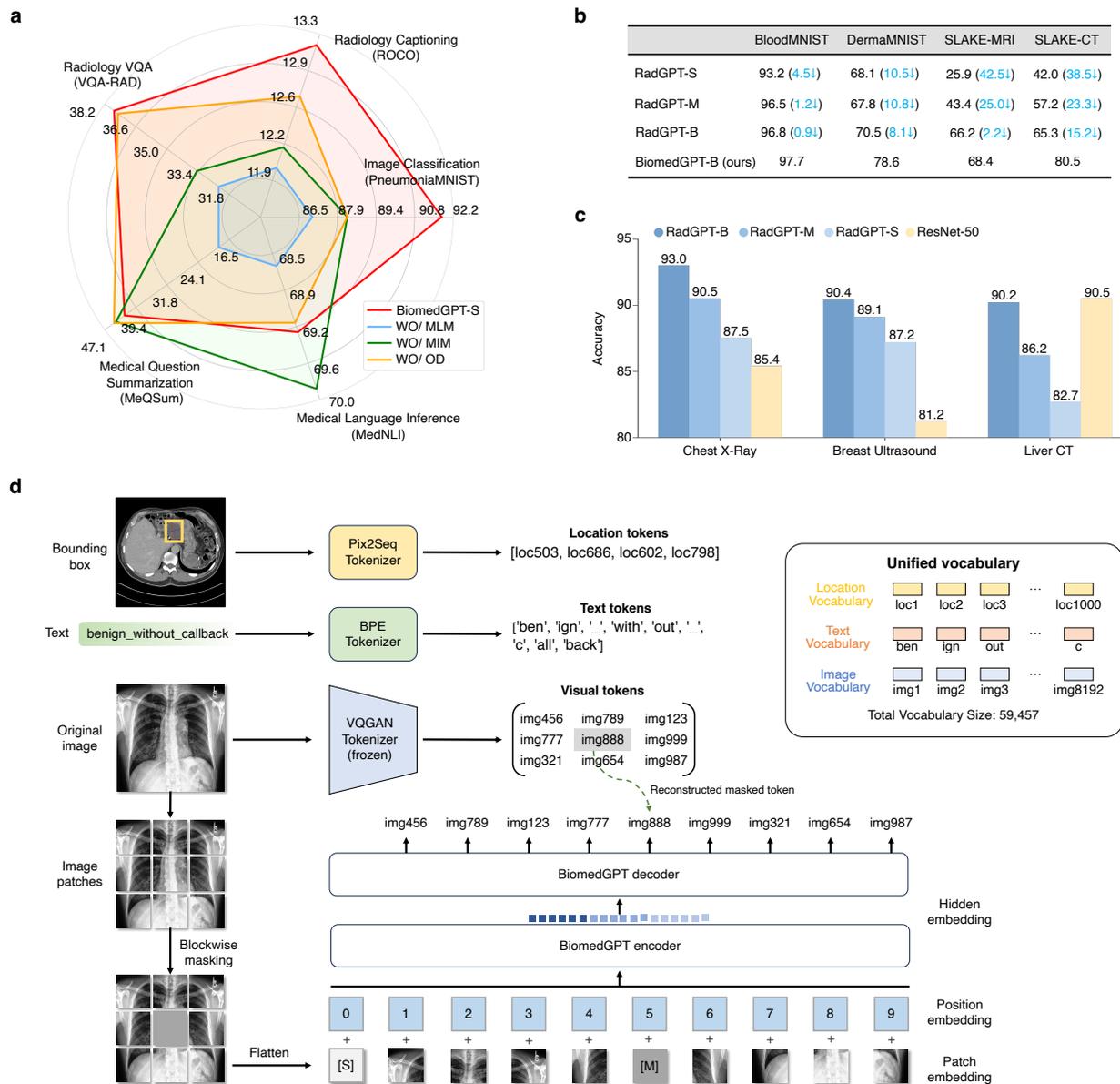

**Fig. 6**: Results of ablation study on the impact of diversity of pre-training datasets and tasks and the graphical demonstration of BiomedGPT's design. **(a)** Performance comparison excluding the specific task. Metrics used here are accuracy for radiology VQA, medical language inference and image classification, CIDEr for radiology captioning, ROUGEL-L for medical question summarization. **(b)** Cross-domain transferability of BiomedGPT across four datasets. RadGPT is a variant of BiomedGPT but was pre-trained with radiology-only data. Besides, SLAKE-MRI and SLAKE-CT are the modality-specific subsets of the SLAKE data. **(c)** In-domain transferability of BiomedGPT across three radiology modalities/datasets. **(d)** Description of the unified vocabulary used in BiomedGPT for pre-training and inference. Tokenization of bounding box and text via Pix2Seq and BPE, respectively. There are three types of tokens: location tokens, text tokens and image tokens from the frozen pre-trained tokenizers such as VQ-GAN. We also show the graphical illustration of masked image modeling in the pre-training, which aims to learn the representations via reconstructing the masked patches.

# Methods

Our proposed BiomedGPT is a transformer-based architecture specifically designed for the biomedical field, built upon the success of existing unified models for general data. We follow the fundamental principles of a unified model [52]: 1) modality-agnostic, 2) task-agnostic, and 3) modality and task comprehensiveness. By discretizing data into patches or tokens, we achieve input/output unification using ideas from ViT [59] and language models [10, 11].

## BiomedGPT architecture

There are three principal architectures among pre-trained foundation models: encoder-only, decoder-only, and encoder-decoder. Encoder-only models, such as BERT and its variants [60], primarily utilize the transformer's encoder to learn representations of input data, requiring additional modules like classification heads or task-specific decoders during fine-tuning. This architecture may struggle with aligning inputs and outputs across distinctly different modalities, limiting its capability in complex zero-shot prediction or generation tasks. Conversely, decoder-only models, exemplified by GPT [10], rely solely on the transformer's decoder to process raw text inputs. Although proficient in text-based tasks, their architecture is not inherently equipped to handle multiple modalities, often leading to challenges in learning joint representations across diverse data types. This can diminish flexibility and performance in multi-modal tasks, particularly in biomedical applications. Therefore, we advocate for the encoder-decoder architecture for BiomedGPT, which is more adept at mapping various modalities into a unified semantic representation space, thereby enhancing task handling across a broader spectrum.

BiomedGPT is implemented with a BERT-style encoder [60] over corrupted text and a GPT-style left-to-right autoregressive decoder [10]. All these models rely on the transformer with the multi-head attention mechanism (**Extended Data Fig. 3a**) that is popularly used which allows the model to jointly attend to the information from different representation sub-spaces [61]. To improve the convergence efficiency and stability in the pre-training, we add three normalization operations to each layer: a post-attention Layer Norm (LN) [62], post-first-FFN LN, and head-wise scaling within self-attention (**Extended Data Fig. 2b**), following [63]. To encode positional information, we incorporate two sets of absolute position embeddings for both text and images. Rather than merely combining these embeddings with token and patch embeddings, we implement a decoupling method to separate position correlation (**Extended Data Fig. 3b**), which may bring unnecessary randomness in the attention and further limit the expressiveness of the model [61]. Furthermore, we also incorporate 1D relative position bias for text and 2D relative position bias for image (**Extended Data Fig. 3c**), as described in previous works [64, 65]. To investigate the performance of BiomedGPT for tasks at different scales, we explicitly designed three scaling models, i.e., BiomedGPT-S (33M), BiomedGPT-M (93M), and BiomedGPT-B (182M). The configurations for each model are detailed in **Extended Data Fig. 2a**.

## Unifying input/output

To handle diverse modalities without relying on task-specific output structures, we represent them with tokens drawn from a unified and finite vocabulary (**Fig. 6d**). To achieve this, we utilize the frozen image quantization [66] and object descriptor [67] to discretize the images and objects on the target side, respectively. For text outputs, including object labels and summarizations, we encode them using BPE tokens [68]. Specifically, an image with a resolution of 256 × 256 is sparsely encoded into a sequence of 16 × 16, which strongly correlates with the corresponding patch and can effectively reduce the sequence length of the image representation. The bounding boxes of objects in an image are expressed as sequences of location tokens in the format of integers. We hereby build a unified vocabulary for all tokens of multimodal outputs. The total vocabulary size is 59457, with 50265 language tokens, 1000 location tokens, and 8192 vision tokens. Note that the number of vision tokens is determined by the variant of the pre-trained VQ-GAN models used in BiomedGPT, specifically, we used the variant with the patch size of 8 and the vocabulary size of 8192. During training, we randomly subsample 196 image patches for pre-training. The truncation to max model input length is set as 512.

**Ablation study on modality comprehensiveness.** Additional evaluations were conducted to address the query: *"Can the proposed model handle unseen data modalities (e.g., images from a new different imaging device like an ultrasound)?"* To investigate this, we adjusted our dataset selection for both pre-training and downstream tasks (**Supplementary Fig. 2b**). Specifically, we used all 3,489 and 6,461 chest X-ray image-text pairs from SLAKE and IU X-ray datasets, respectively. Additionally, we randomly selected 7,452 images from CheXpert while disabling masked language modeling and object detection during pre-training for simplification (**Supplementary Fig. 2a**). The pre-trained BiomedGPT on X-ray modality, denoted as RadGPT-*{size}*), is subsequently fine-tuned on radiology-related datasets: chest X-ray, breast ultrasound and liver CT (coronal-view). As a comparative baseline, we selected ResNet-50 [69], which was trained from scratch on these three datasets. We observed an impressive in-domain transferability of BiomedGPT from the outcome (**Fig. 6c**), specifically, RadGPT-B outperformed the baseline, achieving 93.0% classification accuracy on the chest X-ray images, gaining 7.6% improvement. However, for liver CT scans, we observed a necessity to scale up the model to attain comparable results to the baseline. This highlights the challenges in domain adaptation for medical applications when the pre-trained model does not learn diverse medical knowledge.

We further explored the aspect of cross-domain transferability (**Fig. 6b**). Specifically, we fine-tuned the aforementioned pre-trained model, RadGPT, using datasets from other domains such as blood cell microscopy and dermoscopy for image classification. Additionally, we selected MRI-only and CT-only image-text pairs from SLAKE and conducted VQA fine-tuning. The results compared to the benchmark (the original BiomedGPT-B pre-trained with all modalities), were measured in terms of accuracy. We found that cross-modality transfer with our model is feasible, albeit with potentially significant performance degradation. For example, RadGPT-B exhibited a notable decrease in accuracy compared to the baseline on both the DermaMNIST dataset (dermoscopy), with an 8.1% drop, and the SLAKE-CT VQA dataset, with a more significant reduction of 15.2%. It's noteworthy that we had to double the training epochs as compared to the previous fine-tuning with a pre-trained model encompassing all modalities (100 vs. 50). Therefore, we conclude that modality comprehensiveness is critical for a generalist biomedical AI model to facilitate efficient knowledge transfer.

**Natural language as a task instructor**

Multitasking is a key attribute of a unified and generalist model. Following previous literature on language models using prompt/instruction learning [10, 70, 71], and the existing unified frameworks to eliminate task-specific modules, we specify each task with a handcrafted instruction excluding VQA, which are fully specified by their text inputs. BiomedGPT supports abstractions of several tasks, including vision-only, text-only, and vision-language, to achieve task comprehensiveness. We provide details of the pre-training tasks, fine-tuning/inference tasks, as well as their corresponding instructions in the following.

**Pre-training tasks.** We consider two vision-only tasks in the pre-training: for masked image modeling (MIM) as well as image infilling, we borrow the idea of block-wise masking [72] and let the model recover the masked patches in the middle part by generating the corresponding codes (see **Fig. 6d**). The corresponding instruction is *"What is the image in the middle part?"*. For object detection, the model learns to generate the bounding box of an object with the instruction *"What are the objects in the image?"*. As to the text-only task, we adopt the commonly used masked language modeling (MLM), whose logic is similar to the masked image modeling, while the instruction is *"What is the complete text of '{**Text**}'?"*. Two types of multimodal tasks are selected, including image captioning with the instruction of *"What does the image describe?"* and VQA with the instruction of *'{**Question**}'*. The addition of object detection (OD) for pre-training BiomedGPT serves to enhance visual learning inspired by [73]. The mixture of pre-training tasks is demonstrated to be effective, especially for processing multimodal inputs (**Fig. 6a**).

**Fine-tuning and downstream tasks.** Besides image captioning and VQA used in pre-training, we cover one more vision-only task and two more text-only tasks. Specifically, we use the instruction *"What does the image describe?"* to differentiate image classification. *"What is the summary of text '{**Text**}'?"* and *"Can text1 '{**Text1**}' imply text2 '{**Text2**}'?"* are exploited for text summarization and natural language inference, respectively. It is noteworthy that BiomedGPT is extendable, allowing for customization of instructions for specific downstream tasks (**Fig. 1c and Supplementary Figs. 4-9**).

**Ablation study on task comprehensiveness.** To gain a deeper understanding of the impact of individual pre-training tasks on downstream performance, we implemented an ablation study that excludes either image-only or text-only tasks during pre-training and subsequently fine-tuning the resultant models on five downstream tasks. To ensure a fair comparison, we utilized downstream datasets that were excluded from the pre-training phase: (1) PneumoniaMNIST [36] for image classification; (2) ROCO (https://github.com/razorx89/roco-dataset) for image captioning; (3) VQA-RAD for VQA; (4) MeQSum for text summarization; (5) MedNLI for text understanding. Moreover, each model was fine-tuned using consistent training receipts across the same datasets.

Due to the limited computing resources, we performed this study using BiomedGPT-S only. For short, we denote the pre-training without using masked image modeling, without using masked language modeling, and without using object detection as w/o MIM and w/o MLM, respectively. Referring to **Supplementary Fig. 2c**, we used the BiomedGPT-S model, pre-trained with all tasks, as the baseline. We observed several empirical phenomena in this ablation study (**Fig. 6a**): (1) Excluding the MIM component resulted in decreased performance in image-centric and multimodal tasks, such as image classification and VQA accuracy. Conversely, text-centric tasks showed improvement. These outcomes indicate that MIM is not crucial for text-only tasks, potentially explaining the enhancements in those areas. (2) When MLM was excluded during pre-training, performance declined across all tasks in downstream evaluation. Text-centric tasks were significantly impacted. These findings underscore the importance of MLM for unified models, even for image-only tasks that require text-token dictionaries for label generation. (3) Excluding object detection during pre-training led to notable performance reductions in tasks such as image classification and radiology captioning. However, changes in performance for other datasets were relatively minor, likely due to the limited number of object detection samples and the weak connection to language-only tasks. In summary, our study highlights the importance of task diversity in pre-training for the unified medical AI. While the exclusion of image-specific tasks might benefit performance on text-only downstream, a varied task regime is vital for maintaining generalization across both unimodal and multimodal applications.

**Model pre-training**

We adopted sequence-to-sequence (seq2seq) learning [74], which is the commonly used approach for large language models, to train our BiomedGPT. Formally, suppose we are given a sequence of tokens $\mathbf{x}_{i,b}$ as input, where $i = 1, \cdots, I$ indexes the tokens in a data sample and $b = 1, \cdots, B$ indexes a sample in a training batch. Let a model be parametrized by $\theta$. Then we autoregressively train the model by minimizing:

$$L_\theta(\mathbf{x}_{1,1}, \cdots, \mathbf{x}_{i,b}) = -\sum_{b=1}^{B} \log \prod_{i=1}^{I} p_\theta(\mathbf{x}_{i,b} | \mathbf{x}_{1,b}, \cdots, \mathbf{x}_{i-1,b}) = -\sum_{b=1}^{B} \sum_{i=1}^{I} \log p_\theta(\mathbf{x}_{i,b} | \mathbf{x}_{<1,b}).$$

In the context of BiomedGPT, **x** could refer to both linguistic and visual tokens in the pre-training tasks, including subwords, image codes, and location tokens. Specifically, subwords are extracted by a BPE tokenizer, and we mask 15% of the tokens of the subwords in input in the masked language modeling task as these medical words show relatively high overlapping degrees. For the object detection task, location tokens are generated following Pix2Seq [67] conditioned on the observed pixel inputs. We need data preprocessing for quantizing biomedical images using VQ-GAN [68] because they are surrounded by trivial semantics, e.g., black background and the unmet input size. Therefore, we first remove the trivial background and crop the image to the bounding box of the object of interest, then resize the cropped image to be 256 × 256 and feed the center part with 128 × 128 resolution into the pre-trained VQ-GAN to generate the corresponding sparse image codes, which are the target output in masked image modeling task. Vision-language tasks follow the same tokenization flow. Note that for fine-tuning, we also apply seq2seq learning but with different datasets and tasks.

To pretrain our BiomedGPT, we use the AdamW [75] optimizer with hyperparameters $\beta_1 = 0.9$, $\beta_2 = 0.999$, and $\varepsilon = 1e-8$. The peak learning rate is set to 1e−4, and we apply a linear decay scheduler with a warmup ratio of 0.01 to control the learning rate. For regularization, we set dropout to 0.1 and use a weight decay of 0.01. To enhance the training process, we use stochastic depth with a rate of 0.1 applied to the encoder and decoder, except for convolution blocks. Furthermore, we employ a diversified approach in mixing all pre-training data within each batch. This includes an assortment of multimodal, text-only, vision-only, and object detection samples.

The ratio applied is 8:2:1:1, which emphasizes learning and enhancing the interaction between vision and language. In addition, to address the potential feature shift caused by the inherent modality imbalance within the pre-training data, we adopted modality sampling strategies in each pre-training batch to ensure balance. The models are pre-trained with 10 NVIDIA A5000 GPUs and mixed precision [76]. Specifically, the base, medium, and small-scale models take approximately 87, 32, and 9 hours, respectively. We initialized BiomedGPT with the pre-trained OFA model [52] and adapted it to the biomedical domain using our curated multimodal biomedical dataset. Specifically, we continued training from OFA's pre-trained checkpoints to align biomedical concepts using diverse modality data through masked modeling, object detection, and image-text matching (**Extended Data Table 3**). This approach could reduce computational efficiency as the continued training incorporates general-domain knowledge from OFA, including language understanding capabilities beneficial for question-answering tasks.

## Model fine-tuning and inference

Fine-tuning, a form of transfer learning, involves adapting a pre-trained model's weights to new data. The practice of fine-tuning pre-trained models, a widely acknowledged and highly effective approach in natural language processing and computer vision, has also found significant application in medical AI [77, 78]. Different from most previous biomedical models that necessitate the addition and training of extra components, such as a linear output layer or a decoder, our BiomedGPT model solely relies on fine-tuning the existing structure. The specific instructions employed for this fine-tuning procedure mirror the pre-training workflow, thereby maintaining consistency and efficiency in model adaptation. We observed that in tasks requiring long-context outputs, such as image captioning, the model's performance is influenced by hyper-parameters, specifically beam search size and output length constraints (**Supplementary Table 6**). These findings informed our selection of hyperparameters for fine-tuning, which should be based on data statistics from the training set, such as the maximum length of the target text (**Supplementary Table 7**). For datasets with an official split, we selected the checkpoint that achieved the highest metric on the validation data for inference during model evaluation (**Supplementary Table 7**). For datasets lacking an official split, we employed K-fold cross-validation, used the checkpoint from the last epoch for inference, and reported the mean and standard deviation.

Similar to existing large language models and multimodal models [28], in inference, we used decoding strategies such as beam search to improve generation quality. However, such an approach poses challenges for classification tasks, including unnecessary search of the entire vocabulary and the possibility of generating invalid labels beyond the closed label set. To tackle these issues, we apply a beam search strategy that incorporates a prefix tree (also known as a trie), limiting the number of candidate tokens and resulting in more efficient and accurate decoding. **Extended Data Fig. 3d** demonstrates an example of trie-based beam search; along the path across "Lipid" and "breakdown", BiomedGPT sets logits for all invalid tokens ("mechanism" and "pathway") to $-\infty$ when computing log-probabilities for the target token "in". It is worth noting that trie-based search is also applied during the validation phase of the fine-tuning stage for acceleration (approximately 16× increase in speed in our experiments).

## Model instruction-tuning and zero-shot prediction

Instruction-tuning was developed to improve the question understanding capabilities of the pre-trained BiomedGPT. Following the data curation method of LLaVA-Med [12], we diverged from the traditional VQA approach which typically employs a pre-built answer set during both training and inference. Instead, our instruction-tuning method utilizes an open-vocabulary setting, allowing the model to operate without a predefined set of answers, thereby enabling it to independently determine the most appropriate response during both the training and inference phases.

We summarized experimental settings for each zero-shot trial as follows. In the VQA-RAD zero-shot experiment (**Fig. 4**), we used the original questions from the dataset as prompts or instructions. For the disease diagnosis zero-shot experiments (**Extended Data Fig. 5b**), we employed a common prompt template: "*Does the patient have <disease> given the image?"*. The evaluation datasets were curated based on RSNA Pneumonia Detection Challenge (2018) (https://www.rsna.org/rsnai/ai-image-challenge/rsna-pneumonia-detection-challenge-2018) and MedMNIST v2 (224 × 224 images) [36]. Specific evaluations were conducted across different medical datasets: (1) Pneumonia detection involved 1,000 randomly sampled cases from RSNA, including 548 pneumonia and 452 normal cases. (2) Malignant tumor detection used the BreastMNIST dataset, comprising 114 normal or benign cases and 42 malignant cases. (3) Melanoma recognition was based on a subset of DermaMNIST with 223 positive melanoma cases. (4) Drusen recognition utilized a subset of OCTMNIST, featuring 250 positive drusen cases. (5) Cancer tissue identification was assessed on a PathMNIST subset, which included 1,233 colorectal adenocarcinoma epithelium cases, 421 cancer-associated stroma cases, 339 debris cases, and 741 normal colon mucosa cases. In tuberculosis (TB) detection and report generation using two-view chest X-rays (CXR) (**Extended Data Fig. 5c**), we replicated the experimental settings and prompt templates used by Med-PaLM M. Additionally, we incorporated the MIMIC-CXR training set, which includes single-view image-caption pairs, during continual pre-training to ensure a fair comparison with Med-PaLM M. For report generation, we utilized common NLP metrics to align with Med-PaLM M.

Furthermore, we conducted preliminary zero-shot studies on two instruction-tuned large language models, aiming to explore the upper bounds of in-context learning performance using advanced language backbones. We considered the potential integration of these elements into BiomedGPT to enhance reasoning capabilities. However, these models exhibited significant discrepancies when compared to fine-tuned models (**Supplementary Fig. 3**). These findings suggest that future academic research in medical AI should focus on improving in-context learning abilities and text comprehension, which are crucial for real-world clinical tasks.

## Model extension

BiomedGPT was initially developed to process visual (specifically 2D images) and text data. However, the prototype's capabilities could be extended to encompass additional tasks and modalities. For example, we have extended BiomedGPT to include 3D medical imaging classification (**Extended Data Table 5 and Supplementary Table 4**). This extension involved implementing both pre-training and fine-tuning stages. It only requires integrating a pre-trained 3D VQGAN for tokenizing 3D images in masked image modeling and adding a learnable 3D visual encoder into the pipeline (**Fig. 2a**). To further extend the model's capabilities, especially for non-text generation tasks like segmentation, introducing additional decoders, such as a mask decoder, is appropriate.

## Computing hardware and software

We used Python (version 3.7.4) for all experiments and analyses in the study, which can be replicated using open-source libraries as outlined below. For pre-training, we used ten 24-GB NVIDIA A5000 GPUs configured for multi-GPU training using DistributedDataParallel (DDP) as implemented by the framework PyTorch (version 1.8.1, CUDA 12.2) with the sequence-to-sequence toolkit - fairseq (version 1.0.0). For masked image modeling, we first cropped the middle part of the image and converted it to a sequence of visual tokens based on the pre-trained VQGAN model (https://heibox.uni-heidelberg.de/d/2e5662443a6b4307b470/). Pillow library (version 9.0.1) was used to read images, which were then converted to the base64 string format using Python. Timm library (version 0.6.12), torchvision (version 0.9.1) and opencv-python (version 4.6.0) are applied for image processing and loading during training. We used ftfy library (version 6.0.3) to fix the potential broken Unicode for text processing and loading. Einops library (version 0.6.0) was applied for tensor operations in modeling. For model evaluation, we used pycocotools (version 2.0.4) and pycocoevalcap (version 1.2) to calculate the NLP metrics such as ROUGE-L and CIDEr. Other metrics are calculated based on torchmetrics (version 0.11.0). Numpy (version 1.21.5) and Pandas (version 1.3.5) were used for data collection, pre-processing and data analysis.

## Evaluation metrics

In this paper, we employ various evaluation metrics to thoroughly assess the capabilities of our BiomedGPT model across different tasks. Accuracy is a primary metric used for evaluating the performance in medical image classification, VQA and natural language inference. In addition to accuracy, we also utilize the F1 score for these tasks considering class imbalance, where the F1 score is derived as the harmonic mean of precision and recall:

$$\text{F1} = \frac{2 \times \text{precision} \times \text{recall}}{\text{precision} + \text{recall}}.$$

Specifically, for a more convenient comparison with state-of-the-art approaches, we use the weighted F1 score for VQA. This measure is computed by averaging the F1 scores across each class, with the individual class scores weighted according to their frequency of occurrence:

$$\text{Weighted F1} = \sum_{i=1}^{N} \frac{n_i}{N} \times \text{F1}_i,$$

where $n_i$ is the number of instances in class $i$, N is the total number of instances across all classes, and $\text{F1}_i$ is the F1 score for class $i$. Furthermore, we apply the macro-average F1 score (F1-macro) in image classification tasks on the CBIS- DDSM dataset. The F1-macro score is calculated by determining the F1 score for each class independently and then averaging these scores across all classes. This approach does not account for class imbalances, treating each class with equal importance:

$$\text{F1-macro} = \frac{1}{N} \times \sum_{i=1}^{N} \text{F1}_i.$$

The higher accuracy and F1 score (either weighted- or maro-average), the better performance the model achieves.

ROUGE-L [26] was used to evaluate the quality of the generated text on the tasks of image captioning and text summarization, which stands for recall-oriented understudy for gisting evaluation with the longest common subsequence. Given the candidate $C$ and reference $R$, let $LCS(C,R)$ be the length of the longest common subsequence, which is determined by using dynamic programming, it can be an expression as:

$$\text{ROUGE-L} = \frac{(1 + \beta^2)\, R_{LCS}\, P_{LCS}}{R_{LCS} + \beta^2\, P_{LCS}},$$

Where $R_{LCS} = \frac{LCS(C,R)}{c}$, $R_{LCS} = \frac{LCS(C,R)}{c}$, $\beta = \frac{P_{LCS}}{R_{LCS}}$. $c$ and $r$ represent the length of the candidate and reference. A higher ROUGE-L score means that the generated text shares more of the same sequences of words as the reference text, which typically indicates better quality in terms of capturing the salient points of the reference. It suggests that the generated text is more similar to the reference summaries that it is being compared to, which is usually desired in summarization tasks.

In addition to ROUGEL-L, we also applied METEOR [27] and CIDEr [28] to get a more comprehensive evaluation of captioning generation quality. In detail, METEOR stands for metric for evaluation of translation with explicit ordering. We represented precision and recall as $P = \frac{m}{c}$ and $R = \frac{m}{r}$ and let $m$ be the number of common words in the candidate $C$ and the reference $R$ with the number of words of $c$ and $r$, respectively. The METEOR is calculated via:

$$\text{METEOR} = (1 - p)\frac{PR}{\alpha P + (1 - \alpha)R},$$

where $p$ is the penalty factor and is denoted as $p = \gamma(\frac{ch}{m})^\theta$, $ch$ is the number of chunks, which means a m contiguous ordered block. $\alpha, \theta, \gamma$ are hyperparameters determined according to different datasets.

CIDEr is specifically designed to evaluate the quality of captions of images, which stands for consensus-based image description evaluation. The CIDEr score is calculated based on $n$-gram matching, considering both precision (how many $n$-grams in the generated caption are also in the reference captions) and recall (how many $n$-grams in the reference captions are also in the generated caption). It also weighs the $n$-grams based on their saliency (importance in describing the image) and rarity (uncommonness in the dataset), which helps to emphasize the importance of capturing the most relevant aspects of the image in the caption. Let $c$ be a candidate caption, $S$ be a set of reference captions, and CIDEr is obtained by averaging the similarity of different lengths:

$$\text{CIDEr}_n(c, S) = \frac{1}{M} \sum_{i=1}^{M} \frac{g^n(c) \times g^n(S_i)}{\|g^n(c)\| \times \|g^n(S_i)\|},$$

where $M$ denotes the number of reference captions and $g^n(\cdot)$ denotes an n-gram-based TF-IDF vector. A higher CIDEr score suggests that the generated caption is more accurate and descriptive of the image content, aligning well with human judgments of what the image represents. CIDEr can technically range from 0 to 100. Typically, human captions would tend to score near 90 [28].

# Data Availability

All data in this study are publicly available and can be accessed from: IU X-ray and PEIR GROSS (https://github.com/nlpaueb/bioCaption), MedICat (https://github.com/allenai/medicat), PathVQA (https://github.com/UCSD-AI4H/ PathVQA), SLAKE 1.0 (https://www.med-vqa.com/slake/), DeepLesion (https://nihcc.app.box.com/v/DeepLesion), OIA-DDR (https://github.com/nkicsl/OIA), CheXpert- v1.0-small (https://www.kaggle.com/datasets/willarevalo/chexpert-v10-small), CytoImageNet (https://www.kaggle.com/datasets/stanleyhua/cytoimagenet), ISIC 2020 (https://challenge2020.isic-archive.com), Retinal Fundus (https://www.kaggle.com/ c/diabetic-retinopathy-detection), MIMIC-III Clinic Notes (https://paperswithcode.com/dataset/hospital-admission-notes-from-mimic-iii), NCBI BioNLP (https://www.ncbi.nlm.nih.gov/research/bionlp/Data/), PubMed Abstracts are derived from BLUE benchmark (https://github.com/ncbi-nlp/BLUE_ Benchmark), VQA-RAD (https://osf.io/89kps/), CBIS-DDSM (https://www.kaggle.com/datasets/awsaf49/cbis-ddsm-breast-cancer-image-dataset), SZ-CXR and MC-CXR can be requested via the contact on (http://archive.nlm.nih.gov/repos/chestImages.php), MIMIC-CXR (https://physionet.org/content/mimic-cxr-jpg/2.1.0/), MedNLI (https://physionet.org/content/mednli/1.0.0/), TREC2022 (https://www.trec-cds.org/2022.html), SEER (https://seer.cancer.gov), MIMIC-III (https://physionet.org/content/mimiciii/1.4/), HealthcareMagic (https://github.com/UCSD-AI4H/Medical-Dialogue-System), MeQSum (https://huggingface.co/datasets/sumedh/MeQSum), MedMNIST v2 (https://medmnist.com) , ROCO (https://github.com/razorx89/roco-dataset), a randomly sampled subset of RSNA Pneumonia Detection Challenge (2018) used for zero-shot prediction (https://www.rsna.org/rsnai/ai-image-challenge/rsna-pneumonia-detection-challenge-2018). The MedMNIST-Raw is curated based on multiple sources including NCT-CRC-HE-100K (colon pathology) (https://zenodo.org/records/1214456), HAM10000 (dermatoscopy) (https://github.com/ptschandl/HAM10000_dataset), OCT & Chest X-ray (https://data.mendeley.com/datasets/rscbjbr9sj/3), breast ultrasound (https://scholar.cu.edu.eg/Dataset_BUSI.zip), blood cell microscopy (https://data.mendeley.com/datasets/snkd93bnjr/1), Liver Tumor Segmentation Benchmark (LiTS) (https://competitions.codalab.org/competitions/17094). The VQA data for human evaluation are derived from Medical-Diff-VQA (https://physionet.org/content/medical-diff-vqa/1.0.0/), with the exclusion of questions related to differences, as these require a two-image input. Report generation and summarization samples for human evaluations are extracted from MIMIC-CXR. The instruction-following data used in this article is derived from *Pubmed* (https://pubmed.ncbi.nlm.nih.gov) following LLaVA-Med (https://github.com/microsoft/LLaVA-Med/blob/main/download_data.sh) and is combined with training sets from PathVQA and SLAKE. We also provided the table with more details of the major datasets in **Extended Data Table 2**.

# Code Availability

The pre-trained and fine-tuned models as well as source code for training, inference and data preprocessing can be accessed at https://github.com/taokz/BiomedGPT.

# Extended Data

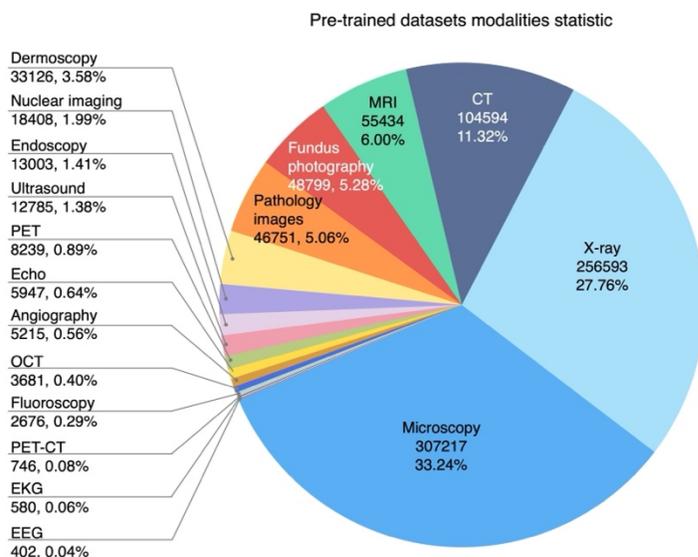

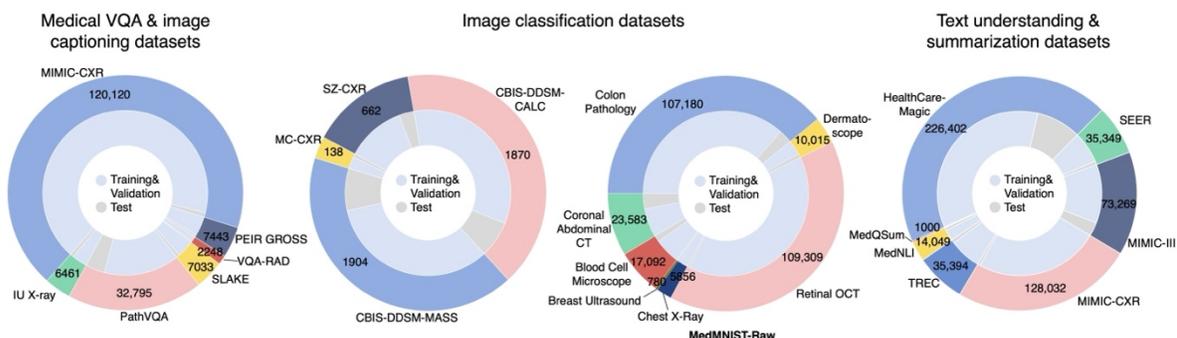

**Extended Data Fig. 1**: Statistics of pre-training and fine-tuning datasets. **(a)** Modality distribution of pre-training data used in BiomedGPT. **(b)** For the training and testing splits of datasets used in downstream fine-tuning, we typically follow the format of number of training samples/number of validation samples/number of test samples to detail each dataset. More details of the data split are described in **Supplementary Table 7**.

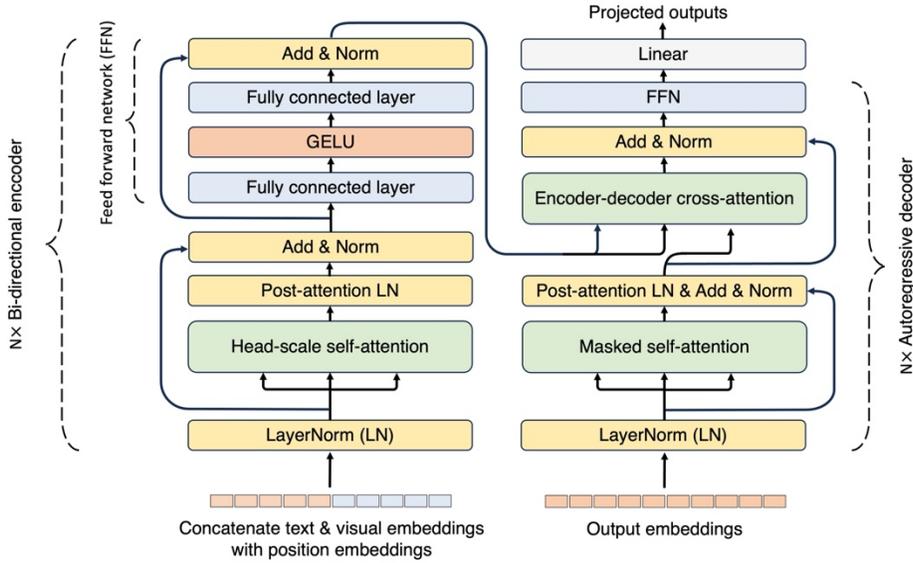

**Extended Data Fig. 2**: Overview of BiomedGPT's model configuration and architecture. **(a)** Detailed model configuration of BiomedGPT. Here, **"#"** indicates *number of*. **"Att."**, **"Enc."** and **"Dec."** indicate *Attention*, *Encoder* and *Decoder*, respectively. The hidden size is the size of the embeddings and the size of the output of each self-attention and feed-forward layer. The first layer of FFN expands the hidden size to the intermediate size, and the second layer contracts it back to the hidden size. This expansion and contraction allow the network to create more complex representations. During the pre-training phase, image processing involves resizing and cropping the images to varying resolutions, corresponding to the input sizes listed in the table. It should be noted that during fine-tuning and inference stages, the input resolution of BiomedGPT can be flexibly adjusted according to the specific requirements of the task. **(b)** The neural network architecture of BiomedGPT, which includes bi-directional encoder blocks and autoregressive decoder blocks. The number of blocks varies for different model scales.

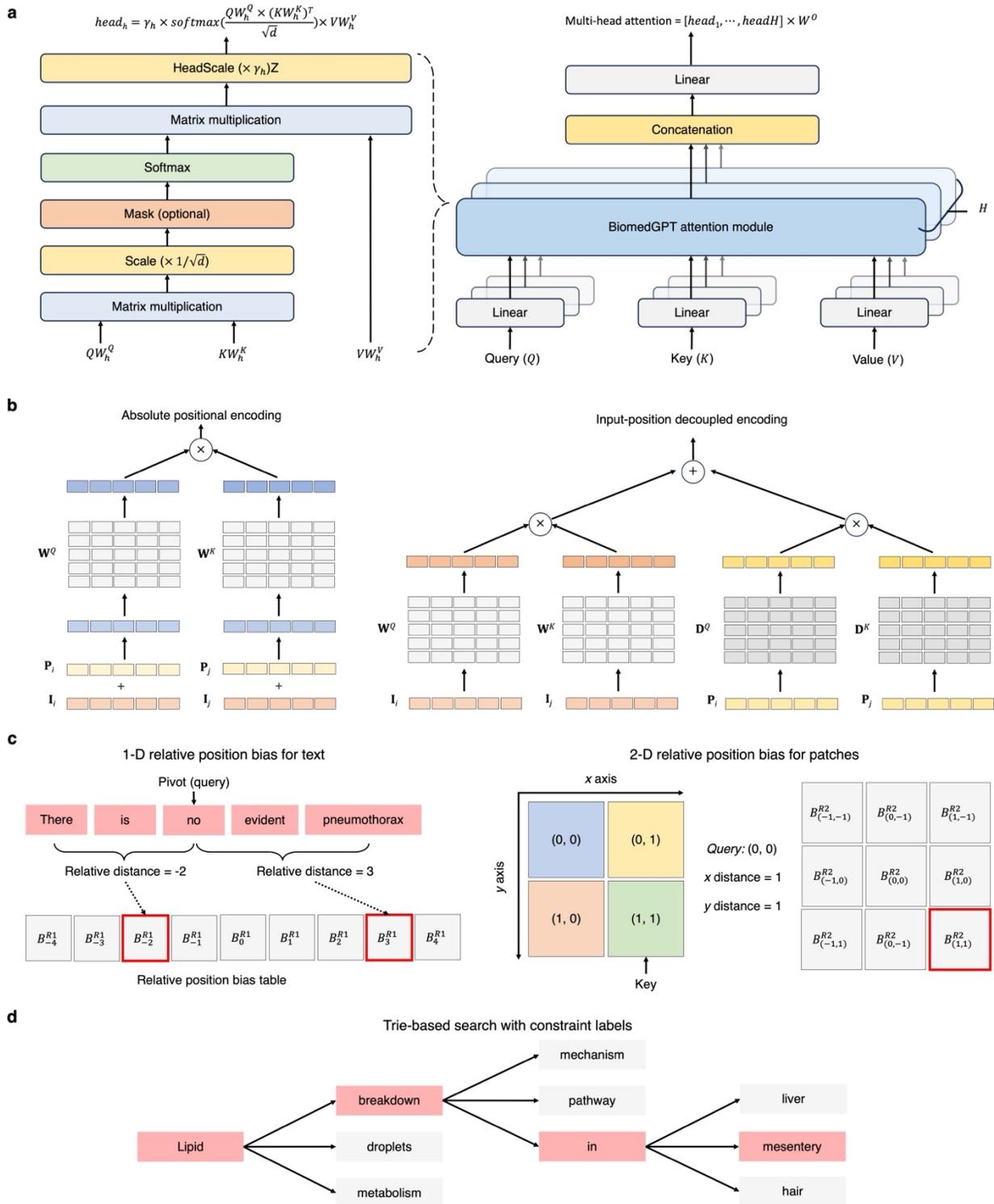

**Extended Data Fig. 3**: The graphical illustrations of the key components in BiomedGPT. **(a)** Head-scale multi-head attention module in BiomedGPT. The trainable parameters $\gamma_h$ is applied prior to the output projection for each head. **(b)** Instead of adding the absolute positional embedding $P_i$ to the input embedding $I_i$ (left), we compute the positional correlation and input correlation separately with different projection matrices and add them together in the self-attention

module (right). **(c)** Graphical illustration of relative position bias. Such an inductive bias $B_{j-i}$ is learnable parameter and can be viewed as the embedding of the relative position $j - i$, which is injected into the Query-Key product: $\frac{1}{\sqrt{d}}(I_i W^Q)(P_i W^K) + B_{j-i}$, and shared in all layers. **(d)** An example of trie-based beam search: along the path across "Lipid" and "breakdown", BiomedGPT sets logits for all invalid tokens ("mechanism" and "pathway") to $-\infty$ when computing log-probabilities for the target token "in". It is worth noting that trie-based search is also applied during the validation phase of the fine-tuning stage for acceleration (approximately 16× increase in speed in our experiments).

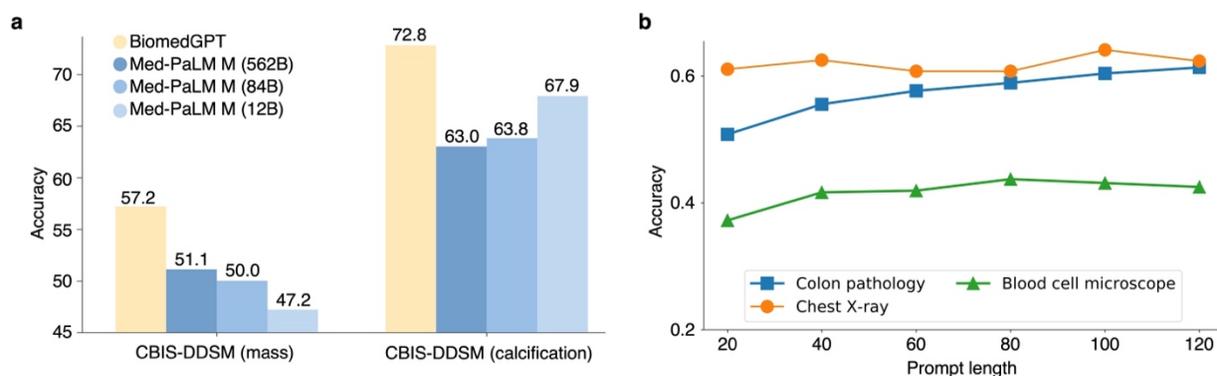

**Extended Data Fig. 4**: Comparative Performance of BiomedGPT and Med-PaLM M and the prompt tuning results in Image classification. **(a)** Comparison between BiomedGPT-B and Med-PaLM M on CBIS-DDSM dataset. **(b)** The experimental results of prompt tuning BiomedGPT-B on three image classification datasets. Prompt tuning learns "soft prompts" or extra model parameters for each task instead of making a task-specific copy of the entire pre-trained model for each downstream task and inference must be performed in separate batches. We must mention that the addition of soft prompts is contrary to the design principle of the generalist model. We injected two prompt layers into the encoder and decoder, and varied the prompt length {20, 40, 60, 80, 100, 120} to investigate the performance comparison against full-model fine-tuning. The preliminary results of "Colon pathology", "Blood cell microscope", and "Chest X-ray" were obtained after 100, 512, and 55 training epochs respectively, all with a consistent batch size of 512. We observed that as the prompt length increases, the model performance tends to improve. However, despite an increased number of tuning epochs compared with fine-tuning on the original BiomedGPT (**Fig. 3c**), the performance after prompt tuning significantly lags behind that of model fine-tuning. Specifically, considering only the best results in prompt tuning, there are substantial accuracy reductions of 32.3%, 54.6%, and 32.6% on these three datasets, respectively.

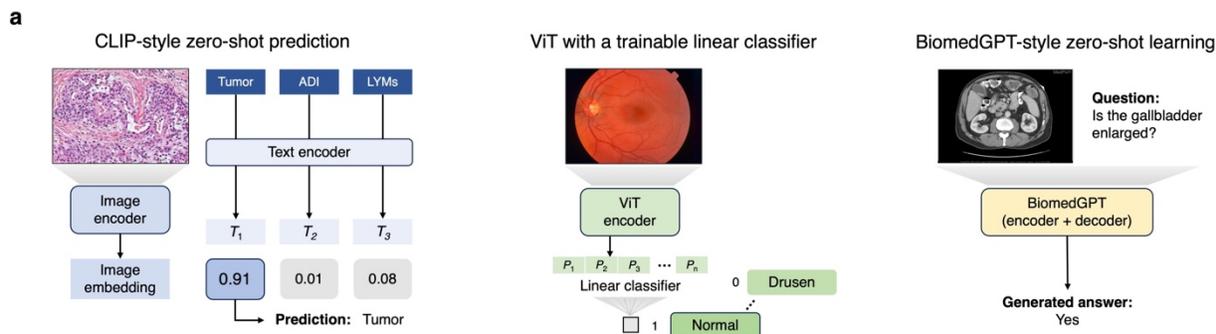

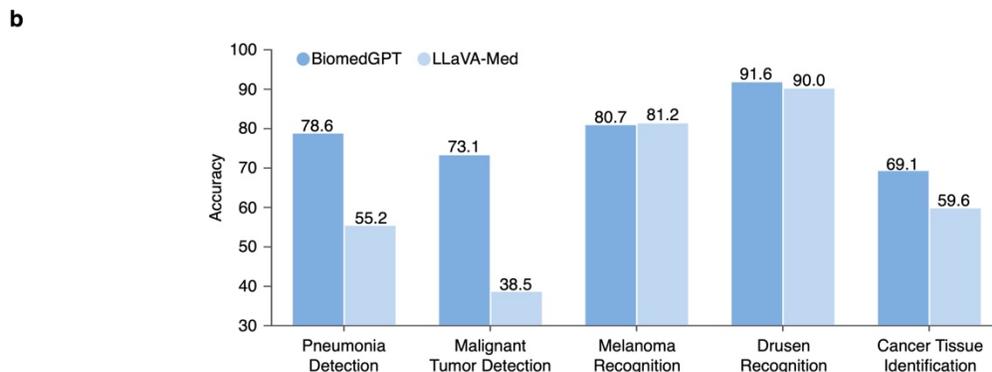

| Task | Metric | Zero-shot | | | | Fine-tuned | |
|---|---|---|---|---|---|---|---|
| | | BiomedGPT (182 M) | LLaVA-Med (8 B) | Med-PaLM M (12 B) | Med-PaLM M (562 B) | BiomedGPT (182 M) | SOTAs |
| TB detection | Accuracy | 78.3 | 34.1 | 87.0 | 87.7 | **89.7** | 88.9 |
| Report generation | ROUGE-L | 24.4 | 17.3 | 27.8 | 28.5 | 28.7 | **29.6** |
| | BLEU-4 | 9.9 | 4.0 | 11.5 | 12.5 | **15.4** | 13.3 |
| | F1-RadGraph | 22.5 | 9.5 | 26.7 | 27.3 | **28.0** | 24.4 |
| | CIDEr | 23.4 | 0.0 | 27.6 | 29.8 | **55.2** | 49.5 |

**Extended Data Fig. 5**: Additional zero-shot results of BiomedGPT. **(a)** Graphical illustration of zero-shot classification using CLIP-style models, linear probing transfer learning using VIT or BERT-style models, and zero-shot generation of BiomedGPT. Notably, our model can generate the response without providing additional components such as the label candidates for CLIP or linear classifier requiring training for ViT. **(b)** Zero-shot performance on five disease diagnosis tasks. **(c)** BiomedGPT shows competitive zero-shot performance compared with Med-PaLM M with a much smaller model scale. The SOTA fine-tuned model for TB detection is TBLightNet. Note that no single model consistently outperforms the others across all four metrics used in report generation. Here, SOTAs represent the best performance achieved in each specific metric. We fine-tuned our pre-trained BiomedGPT-B on MultiMedBench, which Med-PaLM M proposed and used for fine-tuning based on the pre-trained PaLM-E. We also attempted to fine-tune LLaVA-Med; however, the time and computational costs were prohibitive due to the large scale of the model and data. Therefore, we reported the results using the pre-trained checkpoint of LLaVA-Med.

**Extended Data Table 1**: Fine-tuned experimental results of BiomedGPT on 25 diverse experiments.

| Task | Dataset | Domain / Modality | Metric | SOTA Model | SOTA Result | BiomedGPT Small | BiomedGPT Medium | BiomedGPT Base |
|---|---|---|---|---|---|---|---|---|
| Image classification | NCT-CRC-HE-100K | Colon pathology | Accuracy | BiomedCLIP | 91.0 | 94.4 | 95.8 | 95.6 |
| | HAM10000 | Dermatoscopy | Accuracy | MedViT-L | 72.3 | 66.9 | 67.6 | 86.6 |
| | Zhang Lab Data | Retinal OCT | Accuracy | MedViT-L | 89.1 | 84.3 | 92.9 | 90.9 |
| | Zhang Lab Data | Chest X-Ray | Accuracy | BiomedCLIP | 93.0 | 62.5 | 94.9 | 94.9 |
| | Breast Ultrasound | Breast ultrasound | Accuracy | BiomedCLIP | 82.2 | 73.1 | 73.1 | 79.5 |
| | Blood Cell Microscope | Blood cell microscope | Accuracy | BiomedCLIP | 97.9 | 82.7 | 98.5 | 98.7 |
| | LiTS | Coronal abdominal CT | Accuracy | BiomedCLIP | 92.5 | 56.1 | 90.6 | 91.0 |
| | MC-CXR | Chest X-Ray | Accuracy | LightTBNet | 88.9 | 75.9 | 82.8 | 89.7 |
| | SZ-CXR | Chest X-Ray | Accuracy | LightTBNet | 91.0 | 83.5 | 97.0 | 96.2 |
| | CBIS-DDSM | Mass | F1-Macro | Med-PaLM M (562B) | 51.1 | - | 18.7 | 57.2 |
| | CBIS-DDSM | Calcification | F1-Macro | Med-PaLM M (12B) | 67.9 | - | 18.9 | 72.8 |
| Text understanding | MedNLI | Clinic notes | Accuracy | SciFive | 85.6 | 75.8 | 80.8 | 83.8 |
| Clinical-Trial Matching | TREC 2022 | Clinical trials and patient's medical records | Accuracy | LLaVA-Med | 48.7 | 71.6 | 75.4 | 85.2 |
| Treatment suggestions | SEER | Radiation and chemotherapy records | Accuracy | BioGPT | 45.9 | 46.4 | 49.0 | 50.0 |
| Mortality prediction | MIMIC-III | ICU admission notes | Accuracy | UMLS-BERT | 87.3 | 77.8 | 89.2 | 89.0 |
| Text Summarization | MeQSum | Doctor-patient dialogues | ROUGEL-L | BioBART-L | 53.2 | 42.2 | 51.3 | 52.3 |
| | HealthCareMagic | Doctor-patient dialogues | ROUGEL-L | BART-L | 44.7 | 39.8 | 41.99 | 42.0 |
| | MIMC-CXR | Radiology report | ROUGEL-L | RadAdapt | 44.5 | - | - | 44.4 |
| | MIMC-CXR | Radiology report | F1-RadGraph | RadAdapt | 41.8 | - | - | 45.1 |
| | MIMIC-III | Radiology report | ROUGEL-L | MedPaLM M (562B) | 32.0 | - | - | 30.7 |
| | MIMIC-III | Radiology report | F1-RadGraph | MedPaLM M (562B) | 34.7 | - | - | 31.2 |
| Visual question answering | PathVQA | Pathology | Accuracy | CLIP-ViT w/ GPT2 | 63.6 | 47.6 | 49.2 | 58.1 |
| | VQA-RAD | Radiology | Accuracy | MedVInT-TD | 81.6 | 40.1 | 69.4 | 73.2 |
| | SLAKE | Radiology | Accuracy | BiomedCLIP | 85.4 | 69.2 | 81.6 | 86.1 |
| Image captioning | IU X-RAY | Chest X-Ray | CIDEr | PPKED | 35.1 | 29.6 | 31.3 | 40..1 |
| | PEIR GROSS | Digital camera | CIDEr | CoAttention | 32.9 | 22.0 | 25.8 | 122.7 |
| | MIMIC-CXR | Radiology | CIDEr | MedPaLM M (84B) | 26.2 | - | - | 14.7 |

**Extended Data Table 2**: Datasets used in BiomedGPT for pre-training, fine-tuning, evaluation with details.

| Task | | Dataset | Availability | Description |
|---|---|---|---|---|
| Pre-training | Vision & Language | IU X-ray | https://github.com/nlpaueb/bioCaption | A set of chest X-ray images paired with diagnostic reports. |
| | | MediCat | https://github.com/allenai/medicat | A dataset of medical images, captions, and textual references |
| | | PathVQA | https://github.com/UCSD-AI4H/PathVQA | A dataset of question-answer pairs on pathology images. |
| | | PEIR GROSS | https://github.com/nlpaueb/bioCaption | A set of pathology image-caption pairs from PEIR digital library. |
| | | SLAKE | https://www.med-vqa.com/slake/ | An English-Chinese bilingual dataset of question-answer pairs. |
| | Object Detection | DeepLesion | https://nihcc.app.box.com/v/DeepLesion | A dataset with annotated lesions identified on CT images. |
| | | OIA-DDR | https://github.com/nkicsl/OIA | A dataset with annotated fundus images. |
| | Masked Image Modeling | CheXpert | https://www.kaggle.com/datasets/willarevalo/chexpert-v10-small | A set of chest X-ray images with both frontal and lateral views. |
| | | CytoImageNet | https://www.kaggle.com/datasets/stanleyhua/cytoimagenet | A large-scale dataset of openly-sourced and weakly-labeled microscopy images. |
| | | ISIC | https://challenge2020.isic-archive.com/ | Dermoscopic images of unique benign and malignant skin lesions. |
| | | Retinal Fundus | https://www.kaggle.com/c/diabetic-retinopathy-detection | A large set of high-resolution retina images. |
| | Masked Language Modeling | MIMIC-III Clinic Notes | https://paperswithcode.com/dataset/hospital-admission-notes-from-mimic-iii | A dataset of simulated patient admission notes from MIMIC-III. |
| | | NCBI BioNLP | https://www.ncbi.nlm.nih.gov/research/bionlp/Data/ | The corpus contains of annotated PubMed articles. |
| | | PubMed Abstract | https://github.com/ncbi-nlp/BLUE_Benchmark | The corpus consists of annotated PubMed abstracts. |
| Fine-tuning | Medical VQA | PathVQA | https://github.com/UCSD-AI4H/PathVQA | A dataset of question-answer pairs on pathology images. |
| | | SLAKE | https://www.med-vqa.com/slake/ | An English-Chinese bilingual dataset of question-answer pairs. |
| | | VQA-RAD | https://osf.io/89kps/ | A dataset of question-answer pairs on radiology images. |
| | Image Classification | CBIS-DDSM | https://www.kaggle.com/datasets/awsaf49/cbis-ddsm-breast-cancer-image-dataset | A database of scanned film mammography studies. |
| | | MC-CXR | http://archive.nlm.nih.gov/repos/chestImages.php | A dataset of postero-anterior (PA) chest X-rays. |
| | | SZ-CXR | http://archive.nlm.nih.gov/repos/chestImages.php | A dataset of postero-anterior (PA) chest X-rays. |
| | | MedMNIST-Raw | The MedMNIST-Raw is based on multiple datasets:<br>Colon Pathology (NCT-CRC-HE-100K)<br>https://zenodo.org/records/1214456<br>Dermatoscopy (HAM10000)<br>https://github.com/ptschandl/HAM10000_dataset<br>Retinal OCT & Chest X-ray<br>https://data.mendeley.com/datasets/rscbjbr9sj/3<br>Breast Ultrasound<br>https://scholar.cu.edu.eg/Dataset_BUSI.zip<br>Blood Cell Microscopy<br>https://data.mendeley.com/datasets/snkd93bnjr/1<br>Coronal Abdominal CT (LiTS)<br>https://competitions.codalab.org/competitions/17094 | Colon Pathology (NCT-CRC-HE-100K): A set of distinct stained histological images patches.<br>HAM10000: A dataset of dermatoscopic images.<br>Dermatoscopy (HAM10000): A large dataset of labeled OCT and Chest X-ray Images<br>Breast Ultrasound: A dataset of breast ultrasound images.<br>Blood Cell Microscopy: A dataset of microscopic peripheral blood cell images.<br>Coronal Abdominal CT (LiTS): Liver Tumor Segmentation Benchmark. A dataset of enhanced abdominal CT scans. |
| | Image Captioning | IU X-ray | https://github.com/nlpaueb/bioCaption | A set of chest X-ray images paired with diagnostic reports. |
| | | MIMIC-CXR | https://physionet.org/content/mimic-cxr-jpg/2.0.0/ | A database of chest X-ray images with free-text reports. |
| | | PEIR GROSS | https://github.com/nlpaueb/bioCaption | A set of pathology image-caption pairs from PEIR digital library. |
| | | ROCO | https://github.com/razorx89/roco-dataset | A large-scale medical and multimodal imaging dataset. |
| | Text Understanding | MedNLI | https://physionet.org/content/mednli/1.0.0/ | A dataset of sentence pairs created by physicians from MIMIC-III clinical notes. For medical language inference. |
| | | TREC2022 | https://www.trec-cds.org/2022.html | A dataset of physician-curated sentence pairs from MIMIC-III clinical. For clinical trial matching. |
| | | SEER | https://seer.cancer.gov | A dataset includes cancer information and treatment plans for more than 10,000 patients. For treatment suggestion |
| | | MIMIC-III | https://physionet.org/content/mimiciii/1.4/ | A large, de-identified and publicly-available collection of medical records. |
| | Text Summarization | HealthCareMagic | https://github.com/UCSD-AI4H/Medical-Dialogue-System | An English-Chinese bilingual dataset of conversations between doctors and patients. |
| | | MedQSum | https://huggingface.co/datasets/sumedh/MeQSum | A dataset of summarized consumer health questions. |
| | | MIMIC-CXR | https://physionet.org/content/mimic-cxr-jpg/2.0.0/ | A database of chest X-ray images with free-text reports. |
| | | MIMIC-III | https://physionet.org/content/mimiciii/1.4/ | A large, de-identified and publicly-available collection of medical records. |
| Human Evaluation | | Medical-Diff-VQA | https://physionet.org/content/medical-diff-vqa/1.0.0/ | A dataset for difference visual question answering on chest X-ray images. |
| | | MIMIC-III | https://physionet.org/content/mimiciii/1.4/ | A large, de-identified and publicly-available collection of medical records. |
| Instruction-following | | PubMed articles | https://pubmed.ncbi.nlm.nih.gov | PubMed is a free resource supporting the search and retrieval of biomedical and life sciences literature with the aim of improving health–both globally and personally. |
| | | PathVQA | https://github.com/UCSD-AI4H/PathVQA | A dataset of question-answer pairs on pathology images. |
| | | SLAKE | https://www.med-vqa.com/slake/ | An English-Chinese bilingual dataset of question-answer pairs. |

**Extended Data Table 3**: Instructions for pre-training tasks along with the corresponding format of the output. Here, ** represents the image token derived from VQ-GAN's vocabulary. *<loc>* represents the location token. The instruction for the VQA task is the question itself from the dataset.

| Task | Instructions | The example of output |
|---|---|---|
| Masked image modeling | What is the image in the middle part? | <img111> <img222> <img333> ... <img999> |
| Masked language modeling | What is the complete text of "Effect of <mask> on cultured fibroblasts" ? | Effect of **chloroquine** on cultured fibroblasts |
| Object detection | What are the objects in the image? | <loc111> <123> <loc789> <loc567> **chest** <loc222> <333> <loc666> <loc999> **kidney** |
| Image captioning | What does the image describe? | Interval placement of endotracheal tube and nasogastric tube in standard position. |
| Visual question answering | {**Question**} | {**Answer**} |

**Extended Data Table 4**: Description of the question types in the selected VQA-RAD data samples, which are used for the evaluation of zero-shot learning performance.

| Type | Explanation |
|---|---|
| Modality recognition | The specific imaging modality, such as CT, MRI, or others. |
| Structural identification | The specific anatomical landmarks or structures within the captured images. |
| Lesion & abnormality detection | The identification of anomalous patterns or aberrations |
| Disease diagnosis | Specific disease or medical conditions based on imaging manifestations |
| Size & extent assessment | The dimensions and spread of a lesion or abnormality. |
| Spatial relationships | The relative positioning or orientation of imaged structures. |
| Image technical details | The nuances of the imaging process itself, such as contrast utilization or image orientation |

**Extended Data Table 5**: 3D medical image classification performance in terms of accuracy and F1-Macro. (Details of data and training are described in **Supplementary Table 4**).

| Model | Parameters | AIBL | | MIRIAD | | LIDC | |
| --- | --- | --- | --- | --- | --- | --- | --- |
| | | Accuracy | F1-macro | Accuracy | F1-macro | Accuracy | F1-macro |
| BiomedGPT-B-3D | 182 M | 88.6 | 77.8 | 84.7 | 83.0 | 92.9 | 92.1 |
| BiomedGPT-M-3D | 93 M | 84.7 | 72.1 | 80.0 | 77.5 | 89.9 | 88.9 |
| MedicalNet-101 | 99 M | 81.8 | 66.8 | 70.6 | 65.5 | 89.9 | 88.5 |
| MedicalNet-152 | 152 M | 85.7 | 72.6 | 78.2 | 75.9 | 90.9 | 89.5 |
| COVID-ViT | 78 M | 64.4 | 51.0 | 33.8 | 33.3 | 91.9 | 90.8 |
| Uni4Eye | 340 M | 69.7 | 55.8 | 64.7 | 59.5 | 84.9 | 82.8 |

# Supplementary Information

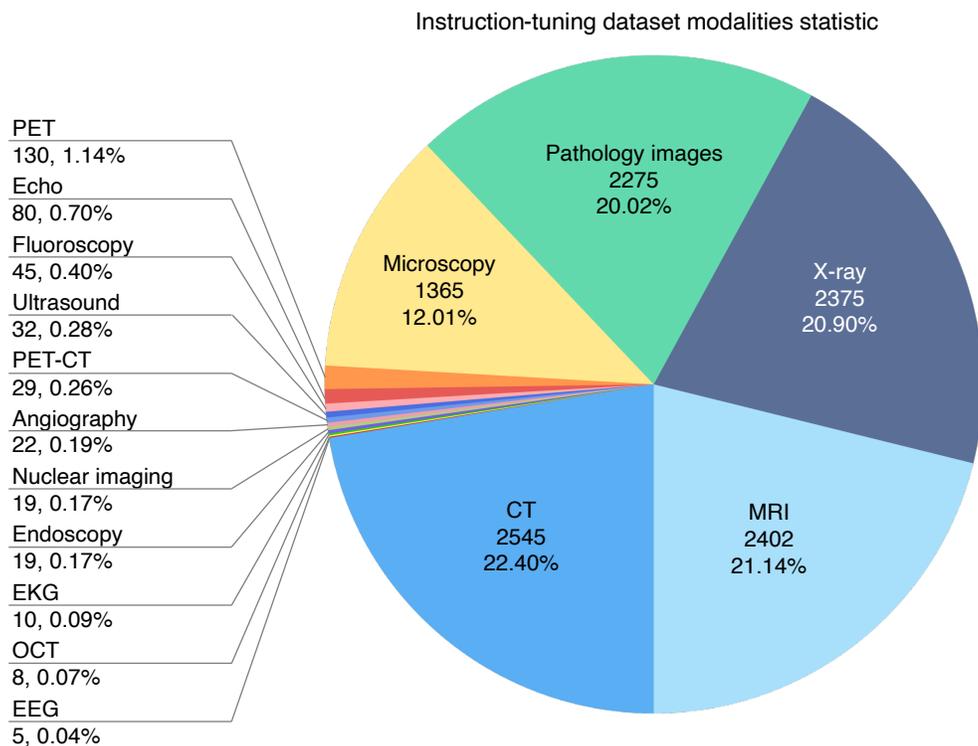

**Supplementary Fig. 1**: Modality distribution of instruction-tuning data used in BiomedGPT.

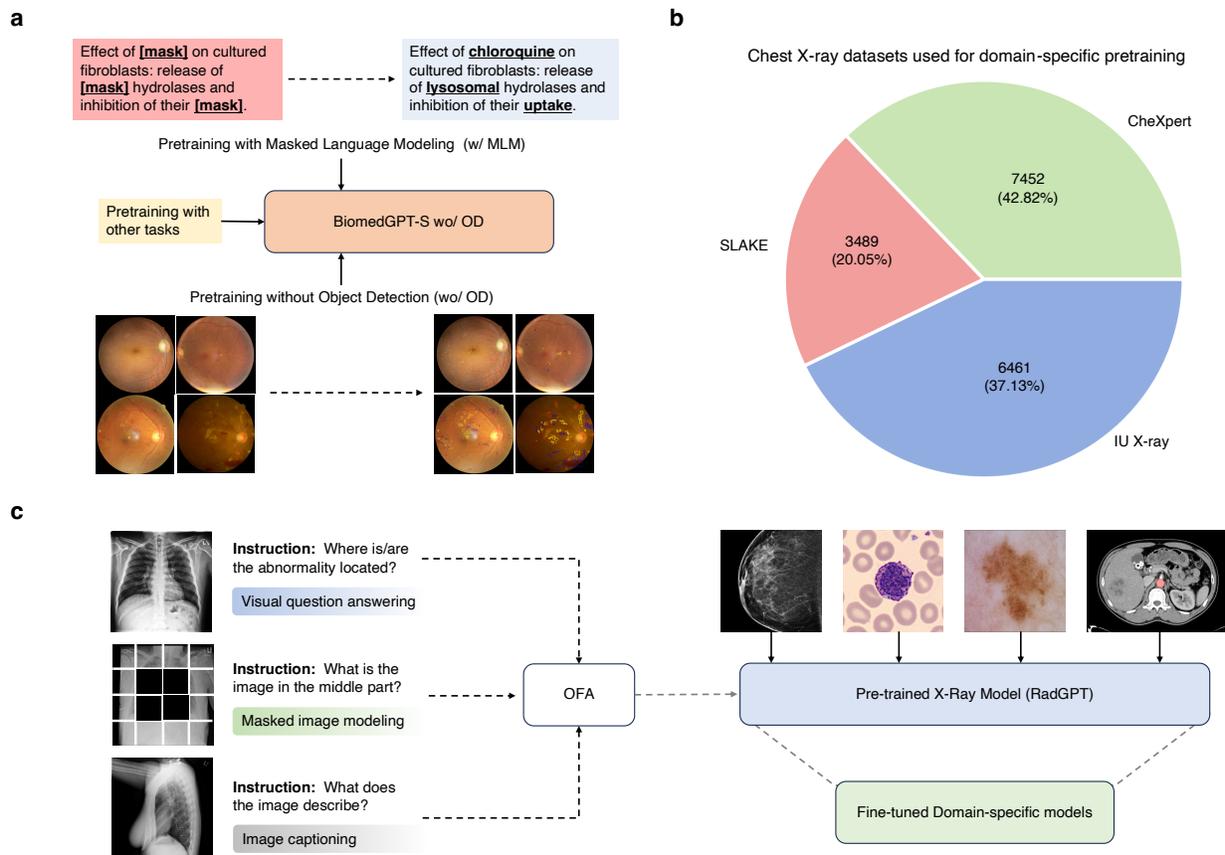

**Supplementary Fig. 2**: Ablation study to demonstrate the impact of diversity of pretraining datasets and tasks. **(a)** Descriptions of exclusion of object detection task and inclusion of MLM with other tasks in the pretraining. **(b)** Distribution of datasets for domain-specific pretraining. **(c)** Description of domain-specific (continual) pretraining with only chest X-ray datasets. The pre-trained model will be further used for fine-tuning on other domains.

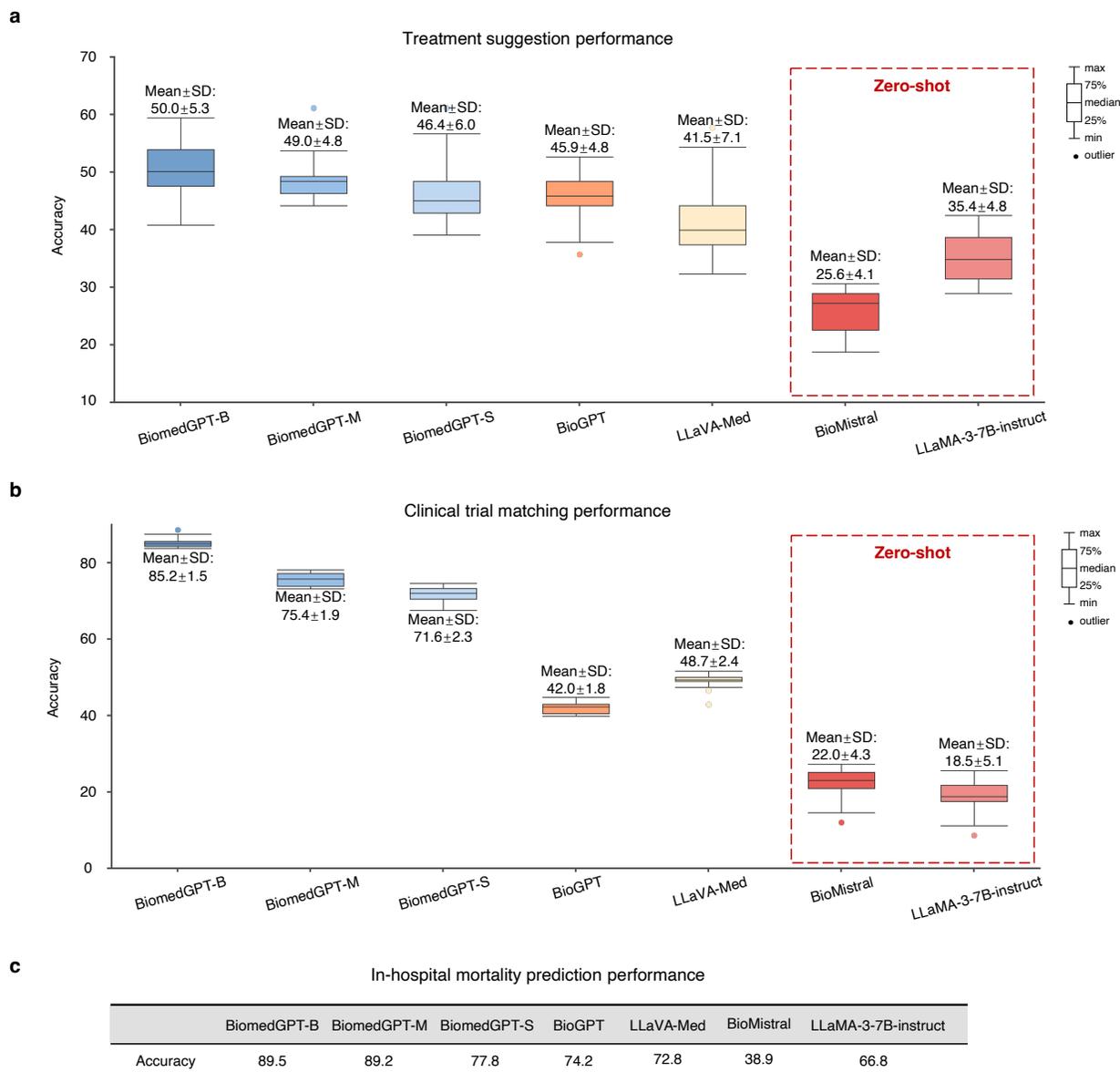

**Supplementary Fig. 3**: Zero-shot performance of BioMistral [1] and LLaMA-3-instruct (https://huggingface.co/meta-llama/Meta-Llama-3-8B-Instruct) was compared with the fine-tuned results of other models on three tasks. **(a)** Treatment suggestion performance ($n$=4680). **(b)** Clinical-trial matching performance ($n$=7079). **(c)** In-hospital mortality prediction performance.

| Image: | 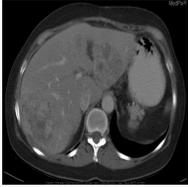 | 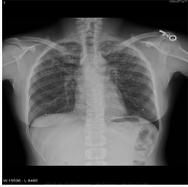 | 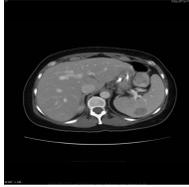 | 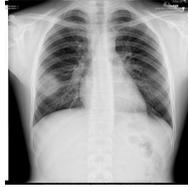 | 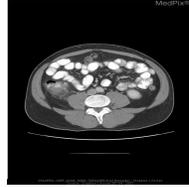 |
|---|---|---|---|---|---|
| Question: | (1) What are the hyperdensities on the periphery of the image? | (2) What is the biological sex of the patient? | (3) Are there multiple or just 1 metastatic focus? | (4) Where is the lesion located? | (5) What is the condition? |
| Answer: | ribs | female | one | right lower lateral lung field | diverticulitis |
| **PubMedCLIP:** | spinal cord | yes | yes | right lower lateral lung field | diverticulitis |
| **BiomedGPT:** | ribs | female | just one | right lower lateral lung field | diverticulitis |

**Supplementary Fig. 4**: Examples from VQA-RAD with PubMedCLIP and our BiomedGPT. It is worth mentioning that, similar to PubMedCLIP [2], previous studies [3, 4] have also struggled to correctly interpret the question, resulting in irrelevant answers for the first three samples. An example is the middle image with the question indexed by (3), where "just one" is correctly inferred. However, the current automatic evaluation framework counts only absolute matches, which may not fully reflect the model's capability to produce semantically accurate answers. As an alternative, we can consider conducting human evaluations and using GPT-4 (https://openai.com/index/gpt-4/) as the judge.

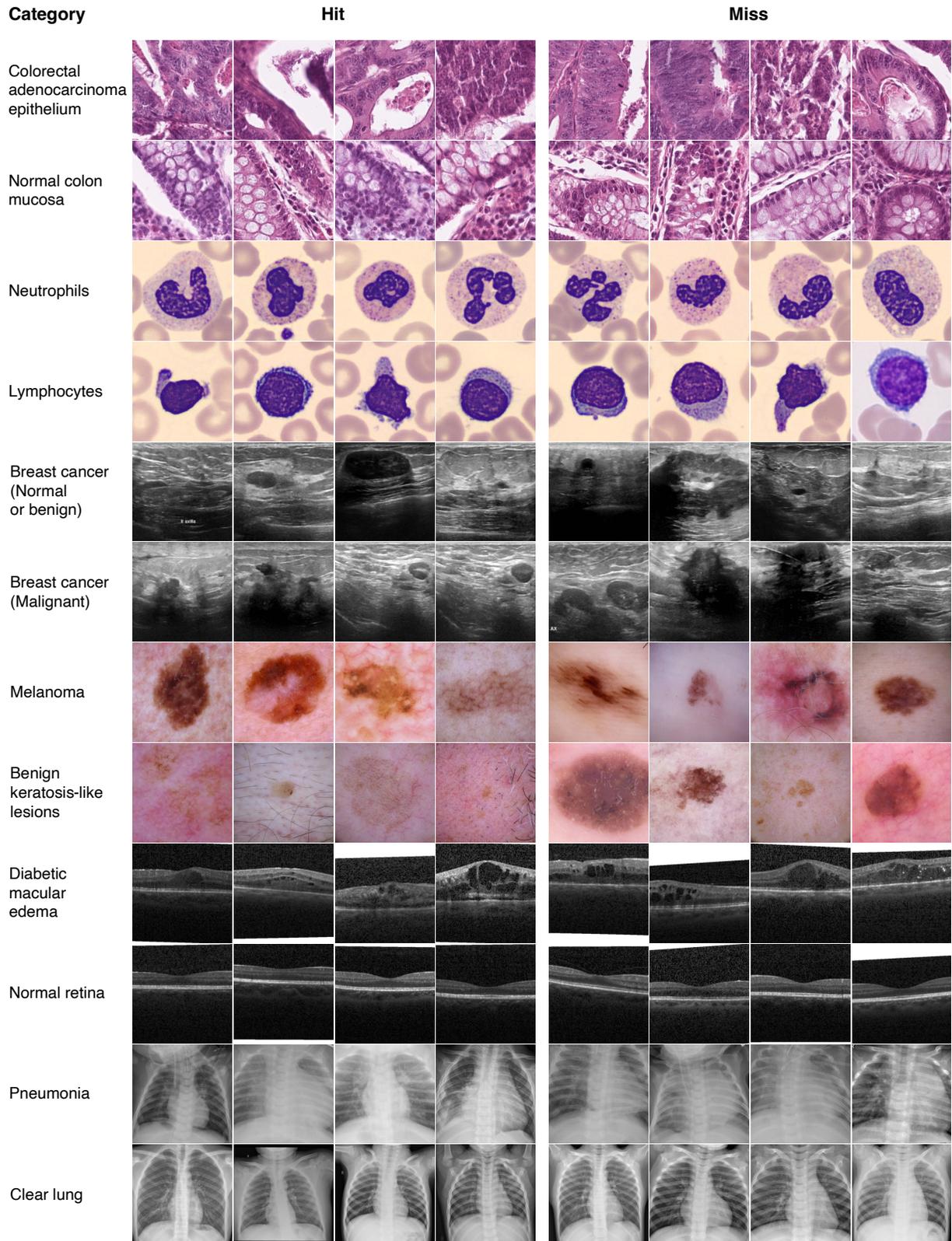

**Supplementary Fig. 5**: Examples of "Hit" (correct predictions by BiomedGPT) and "Miss" (incorrect predictions by BiomedGPT) in disease diagnosis and object recognition.

| Hit: | Hit | Miss: | Significant Error | Insignificant Error | Significant Omission | Insignificant Omission |

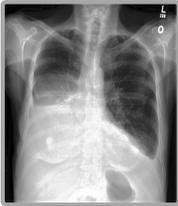

**Reference Report**
There is a right pleural effusion, the size of which is difficult to ascertain. There is unchanged bilateral lower lobe and right middle lobe collapse. The small left pleural effusion is unchanged. There is no pulmonary vascular congestion or pneumothorax. The cardiac and mediastinal contours are not well visualized.

**Generated Report**
A large right pleural effusion is increased from with associated compressive atelectasis. A small left pleural fluid is also increased from __. The left lung is clear. The heart and mediastinum cannot be accurately assessed on this projection. [No mention of changes in lobe collapse or new findings of congestion and pneumothorax.]

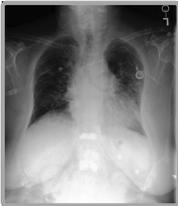

**Reference Report**
A Port-A-Cath terminates in the upper right atrium. The cardiac, mediastinal and hilar contours appear unchanged. Fine reticulation associated with pulmonary fibrosis appears very similar within each lung in extent and distribution with no significant superimposed change. The lung volumes are low. There is no pleural effusion or pneumothorax. Multiple compression deformities including lower thoracic vertebroplasties appear unchanged.

**Generated Report**
Leftsided port terminates in the low S tumult without evidence of pneumothorax. The cardiac and mediastinal silhouettes are stable. Prominence of the interstitial markings bilaterally is stable. No new focal consolidation is seen. No pleural effusion or pneumothorsax is seen. The patient is status post vertebroplasty at the thoracolumbar junction. [No mention of multiple compression deformities].

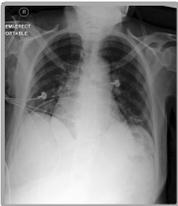

**Reference Report**
The nasogastric tube is in adequate position and there is a resolution of the gastric distention. There is still mild bibasilar atelectasis. There are no pneumothorax no pleural effusion. The cardiac and mediastinal contour are unchanged.

**Generated Report**
The tip of the nasogastric tube is in the body of the stomach. The side port is at the level of the gastroesophageal junction. The tip is not included in this examination. The lung volumes are low with mild bibasilar atelectasis. The heart is mildly enlarged. The mediastinal and hilar contours are normal. The pulmonary vasculature is normal. No pleural effusion or pneumothorax is seen. [No mention of resolution of the gastric distention].

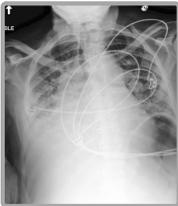

**Reference Report**
There are low lung volumes. The cardiac silhouette size is enlarged, similar when compared to the prior study. There is mild pulmonary edema with perihilar haziness and vascular indistinctness. A small to moderate right pleural effusion is increased when compared to the prior exam. Right basilar opacification may reflect atelectasis, though infection cannot be completely excluded. No pneumothorax is present. Gaseous distention of the stomach noted.

**Generated Report**
The lung volumes are low limiting evaluation. A large right pleural effusion is present with right basilar opacification likely reflecting compressive atelectasis though infection cannot be excluded. The left lung is clear. The heart is probably mildly enlarged. The mediastinal and hilar contours appear unchanged. A mild interstitial abnormality suggests fluid overload.

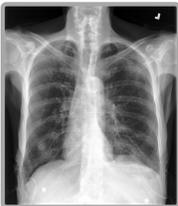

**Reference Report**
There is a mild-to-moderate left pneumothorax with rightward mediastinal shift more apparent than on portable chest radiograph at 2:26 p.m. The small right pneumothorax is stable. There is also a moderate left pleural effusion. Bilateral pigtail catheters are in place. The heart size remains normal. There is no focal consolidation.

**Generated Report**
A leftsided chest tube is in unchanged position. A tiny right apical pneumothorax is unchanged from the prior study. A left pleural drain is in stable position. The cardiomediastinal silhouette is within normal limits. The lungs are clear without focal consolidation or pleural effusion. [No mention of mild-to-moderate left pneumothorax with rightward mediastinal shift].

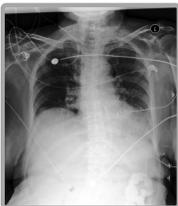

**Reference Report**
Lung volumes are low. Elevation of the right hemidiaphragm appears similar. Cardiomegaly is again noted. Minimal linear left basilar opacity appears similar and likely represents atelectasis. Of note, evaluation is slightly limited in the absence of lateral view. No pleural effusion or pneumothorax is seen on this single view. No focal consolidation is seen on this single view. Aortic calcifications are again noted. Radiopaque material in the left abdomen may represent previously ingested oral contrast.

**Generated Report**
The lung volumes are low. The cardiac mediastinal and hilar contours appear unchanged. The lungs appear clear aside from patchy left basilar opacity suggesting minor atelectasis or scarring. A right internal jugular approach central venous catheter terminates in the lower superior vena cava. The lung bases are partly excluded. [No mention of no evidence for pleural effusion or pneumothorax]. [No mention of cardiomegaly]. [No mention of aortic calcifications]. [Radiopaque material in left abdomen].

**Supplementary Fig. 6**: Examples of generated chest X-ray reports ("findings") by BiomedGPT compared with the reference report. The errors and omissions with the severity levels are conducted by the radiologist and are highlighted in this figure.

| Hit | Miss |
|---|---|
| 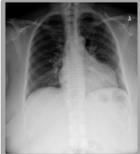 **Q**: What kind of symptoms will the disease on the heart in this image bring to the patient? **A**: Chest tightness, arrhythmia. *Correct* ✓ | 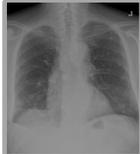 **Q**: What kind of symptoms will the disease on the upper right of lung bring to the patient? **A**: Chest tightness, shortness of breath, difficulty breathing, dry cough. **Reference**: Chest tightness, arrhythmia. *Partially* ✓ *Incorrect* |
| 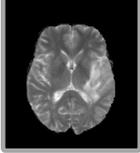 **Q**: How to treat the most severe disease in this image? **A**: Pharmacotherapy, rehabilitation. *Correct* ✓ | 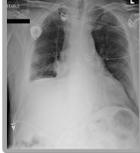 **Q**: What kind of symptoms will the disease on the heart in this image bring to the patient? **A**: Bacterial infection. **Reference**: Bronchial obstruction. *Partially* *Incorrect* ✓ |
| 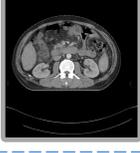 **Q**: What is the function of the organ on the center of this image? **A**: Absorb nutrients, secrete enzymes, digest food. *Correct* ✓ | 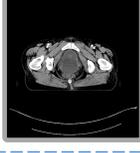 **Q**: What is the effect of the gray organ on the center of this picture? **A**: Store feces, excrete feces. **Reference**: Store urine. *Partially* ✓ *Incorrect* |
| 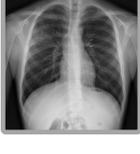 **Q**: How to prevent the occurrence of the disease on the upper right lobe of lung in this image? **A**: Pay attention to prevent cold and keep warm, enhance physical fitness. *Correct* ✓ | 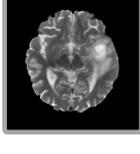 **Q**: What kind of symptoms will the disease on the left lobe of brain in the picture bring to the patient? **A**: Encephalomalacia, local edema, confusion, increased intracranial pressure. **Reference**: Visual impairment, vomiting, tinnitus, increased intracranial pressure. *Partially* ✓ *Incorrect* |
| 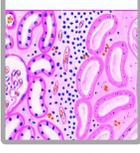 **Q**: What does the interface between viable and non-viable area show? **A**: Non-specific chronic inflammation proliferating vessels. *Correct* ✓ | 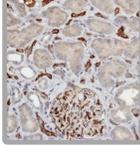 **Q**: What stain shows C4d deposition in peritubular capillaries and a glomerulus? **A**: Immunohistochemical staining. **Reference**: Immunoperoxidase stain. *Partially* ✓ *Incorrect* |
| 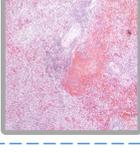 **Q**: Why does this image show spleen, chronic congestion? **A**: Due to portal hypertension from cirrhosis hcv. *Correct* ✓ | 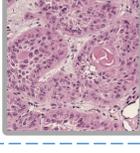 **Q**: How are the tumor cells strikingly similar to normal squamous epithelial cells? **A**: With open chromatin and prominent nucleoli. **Reference**: With intercellular bridges and nests of keratin. *Partially* *Incorrect* ✓ |
| 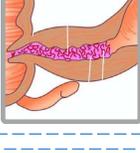 **Q**: What does luminal surface of longitudinal cut section show? **A**: Segment of thickened wall with narrow lumen. *Correct* ✓ | 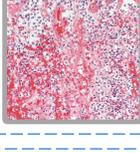 **Q**: What result in acute adrenal insufficiency? **A**: Acute hemorrhagic pancreatitis **Reference**: Bilateral adrenal hemorrhage *Partially* *Incorrect* ✓ |
| 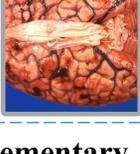 **Q**: Why does this image show brain, subarachnoid hemorrhage? **A**: Due to disseminated intravascular coagulation. *Correct* ✓ | 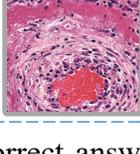 **Q**: What does the wall of the artery show with protein deposition and inflammation? **A**: Pink-red deposits amyloid. **Reference**: A circumferential bright pink area of necrosis. *Partially* *Incorrect* ✓ |

**Supplementary Fig. 7**: Examples of "Hit" (correct answers) and "Miss" (partially correct and incorrect answers) in VQA.

**Findings**

```
Lateral view somewhat limited due to overlying motion
artifact. The lungs are low in volume. There is no
focal airspace consolidation to suggest pneumonia. A
1.2-cm calcified granuloma just below the medial
aspect of the right hemidiaphragm is unchanged from
prior study. No pleural effusions or pulmonary edema.
There is no pneumothorax. The inferior sternotomy
wire is fractured but unchanged. Surgical clips and
vascular markers in the thorax are related to prior
CABG surgery.
```

**Generated impression**

No evidence of acute cardiopulmonary process.

**Reference impression**

No evidence of acute cardiopulmonary process.

---

**Findings**

```
Bedside upright AP radiograph of the chest
demonstrates little interval change when compared to
prior study performed 24 hours ago. There is minimal,
stable enlargement of the cardiomediastinal contours
consistent with mild chronic heart failure.
Persistent obscuration of the pulmonary vascular
markings in the right lung base is consistent with
trace pulmonary edema. Bibasilar atelectasis is still
present. The lungs are otherwise clear. There is no
pneumothorax or pleural effusion. A left internal
jugular central venous catheter, an endotracheal tube,
and an orogastric tube are unchanged and
appropriately positioned. The chronic findings of
atherosclerotic calcification of the aortic arch and
bilateral glenohumeral joint degenerative changes are
once again noted.
```

**Generated impression**

Unchanged mild cardiomegaly and mild pulmonary edema.

**Reference impression**

1. mild chronic congestive heart failure with stable trace pulmonary edema at the right lung base. 2. stable bibasilar atelectasis.

---

**Findings**

```
Single portable view of the chest is compared to
previous exam from ___. Tracheostomy tube is again
noted. Left PICC tip is not clearly delineated on the
current exam. Again there is mild pulmonary vascular
congestion. Streaky opacities at the lung bases
suggestive of atelectasis; however infection cannot
be excluded. Cardiomediastinal silhouette is stable
as are the osseous and soft tissue structures
```

**Generated impression**

No significant interval change. mild pulmonary vascular congestion and bibasilar opacities potentially due to atelectasis, however, infection is not completely excluded.

**Reference impression**

No significant interval change since prior. pulmonary vascular congestion. bibasilar opacities potentially due to atelectasis; however, infection is not excluded.

---

**Findings**

```
Frontal and lateral radiographs of the chest
redemonstrate a round calcified pulmonary nodule in
the posterior right lung base, unchanged from
multiple priors and consistent with prior
granulomatous disease. A known enlarged right hilar
lymph node seen on CT of ___ likely accounts for the
increased opacity at the right hilum. A known right
mediastinal lymph node conglomerate accounts for the
fullness at the right paratracheal region. No pleural
effusion, pneumothorax or focal consolidation is
present. The patient is status post median sternotomy
and CABG with wires intact. The cardiac silhouette is
normal in size. The mediastinal and hilar contours
are unchanged from the preceding radiograph.
```

**Generated impression**

No evidence of acute infection. Unchanged appearance of known right hilar lymph node from prior granulomatous disease.

**Reference impression**

No evidence of acute cardiopulmonary process.

---

**Findings**

```
Left PICC tip is seen terminating in the region of
the distal left brachiocephalic vein. Tracheostomy
tube is in unchanged standard position. The heart is
moderately enlarged. Marked calcification of the
aortic knob is again present. Mild pulmonary vascular
congestion is similar. Bibasilar streaky airspace
opacities are minimally improved. Previously noted
left pleural effusion appears to have resolved. No
pneumothorax is identified. Percutaneous gastrostomy
tube is seen in the left upper quadrant.
```

**Generated impression**

1. Hazy opacity in the right lung could represent aspiration, pleural effusion or hemorrhage.
2. Mild pulmonary edema.
3. Retrocardiac opacity is unchanged.

**Reference impression**

1. Hazy opacity in the right lung which may represent aspiration versus pleural effusion or hemorrhage.
2. Mild pulmonary edema.
3. No displaced rib fractures.

---

**Findings**

```
An endotracheal tube, NG tube, and right upper
extremity PICC with its tip at the cavoatrial
junction are unchanged. There is no change in left
lower lobe opacity. There is no large pleural
effusion, or pneumothorax. The cardiac silhouette
remains moderately enlarged, mediastinal contours are
notable for calcification of the aortic arch.
```

**Generated impression**

No change in left lower lobe opacity, which may be due to atelectasis with or without pleural effusion.

**Reference impression**

Mild residual retrocardiac opacification remains, pneumonia vs. atelectasis.

**Supplementary Fig. 8**: Examples of predictions by BiomedGPT in radiology report generation.

## Hit

**Patient information**
A 19-year-old male came to clinic with some sexual concern. He recently engaged in a relationship and is worried about the satisfaction of his girlfriend. He has a "baby face" according to his girlfriend's statement and he is not as muscular as his classmates. On physical examination, there is some pubic hair and poorly developed secondary sexual characteristics. He is unable to detect coffee smell during the examination, but the visual acuity is normal. Ultrasound reveals the testes volume of 1-2 ml. The hormonal evaluation showed serum testosterone level of 65 ng/dL with low levels of GnRH.
**Clinical trial information**
Subjects who completed ZA-003 were eligible to receive an additional year of treatment in this extension study. The objectives of this study were to evaluate the safety and efficacy of Androxal® administered orally once daily for one year in men with secondary hypogonadism and who had completed ZA-003.
**Inclusion Criteria:** - Total serum testosterone concentrations < 300 ng/dL at baseline
**Exclusion Criteria:** - Presence or history of prostate cancer - Elevated PSA > 3.5 ng/mL Additional inclusion and exclusion criteria may apply.
**Q**: Please provide clinical trial matching result.
**A**: The patient is **eligible** for the clinical trial.

---

**Patient information**
A 32-year-old woman comes to the hospital with vaginal spotting. Her last menstrual period was 10 weeks ago. She has regular menses lasting for 6 days and repeating every 29 days. Medical history is significant for appendectomy and several complicated UTIs. She has multiple male partners, and she is inconsistent with using barrier contraceptives. Vital signs are normal. Serum β-hCG level is 1800 mIU/mL, and a repeat level after 2 days shows an abnormal rise to 2100 mIU/mL. Pelvic ultrasound reveals a thin endometrium with no gestational sac in the uterus.
**Clinical trial information**
**Description:** Missed abortion is a condition where the fetus has perished but the miscarriage is not expelled. Women often present at a routine ultrasound or with a slight brownish discharge. Traditionally this condition has been treated with curettage or vacuum aspiration. Lately, medical treatment has become more common due to less risk of infection and other complications. The routine medical treatment is 800mcg of misoprostol administered vaginally. We wish to examine of repeated doses of 400mcg misoprostol after the initial 800mcg vaginal misoprostol increases efficacy of the treatment.
**Inclusion Criteria:** - over 18 years of age healthy no contraindication to medical treatment visible fetal structures such as yolk sac or fetus
**Exclusion Criteria:** - empty gestational sac more than sparse bleeding or dilated cervix mental instability after diagnosis contraindication to medical treatment.
**Q**: Please provide clinical trial matching result.
**A**: The patient is **ineligible** for the clinical trial.

---

**Patient information**
A 51-year-old man comes to the office complaining of fatigue and some sexual problems including lack of libido. The patient doesn't smoke or use any illicit drug. Blood pressure is 120/80 mm Hg and pulse is 70/min. Oxygen saturation is 99% on room air. BMI is 24 kg/m2. Skin examination shows increased pigmentation. Genotype testing is consistent with homozygosity for the C282Y mutation. Laboratory study shows transferrin saturation of 55% and serum ferritin of 550 μg/L. He is diagnosed as a case of hemochromatosis.
**Clinical trial information**
**Description:** Investigation whether a switch from oral iron to intravenous ferric carboxymaltose can reduce dose requirements of erythropoiesis-stimulating agents (ESA) and improve Hb levels and iron status in adult patients with non-dialysis-dependent CKD who were on a stable ESA/oral iron schedule for 6 months prior to enrolment.
**Inclusion Criteria:** - >18 years of age - Creatinine clearance ≤40 mL/min - Hemoglobin 110-120 g/L - Serum ferritin <100 μg/L or transferrin saturation <20% - Monthly treatment with ESA and oral iron for at least six months before enrolment.
**Exclusion Criteria:** - Other obvious cause of acute or chronic anemia than iron deficiency - Expectation to require hemodialysis within the next six months - Short life expectancy (<1 year) - Pregnancy - Decompensated heart failure - History of allergic reactions to iron preparations and/or anaphylaxis from any cause - Requirement of blood transfusions - Chronic decompensated mental disorder or dementia
**Q**: Please provide clinical trial matching result.
**A**: The patient is **irrelevant** for the clinical trial.

## Miss

**Patient information**
The patient is a 38-year-old man with cough and body ache that started 3 days ago. He had fever and chills at the beginning and has low grade fever at the time of visit. He feels tired and sleepy. His body ache and myalgia get better after using Tylenol. The PCR test for Covid is positive. His vital signs are within normal limits with a body temperature of 37.9 C. There is no lymphadenopathy or white exudates in the pharynx.
**Clinical trial information**
**Description:** Given the high prevalence of COVID19 illness (both SARS-CoV-2 RT-PCR confirmed and highly suspect cases) ···
**Inclusion Criteria:** - Symptomatic with COVID illness (either RT-PCR test confirmed or highly suspect symptoms) - Call into office within the first 7 days of illness - Have any of the following high risk conditions: - Age >60 - HTN, CAD, or chronic heart disease - Diabetes - Chronic kidney disease - Chronic lung disease - Active or recent chemotherapy for malignancy - Organ transplant - Taking Immune-suppressing medications - HIV with CD4 <200 cells/mm3 - Experiencing at least one of the following high risk symptoms: - Severe cough - Fever 100.0F or greater - Diarrhea - Shortness of Breath - Hypoxia
**Exclusion Criteria:** • HCW not meeting high-risk criteria or those with a contraindication to HCQ will not be offered HCQ
**Q**: Please provide clinical trial matching result.
**A**: The patient is **ineligible** for the clinical trial. ✗
**Reference**: The patient is **eligible** for the clinical trial.

---

**Patient information**
A 19-year-old girl comes to the clinic due to a left wrist mass. She noticed swelling on the top of her wrist about 4 months ago and came to the clinic due to cosmetic concerns. Examination shows a nontender, rounded mass on the dorsal wrist that transilluminates with a penlight. Vital signs are normal. The patient needs to type on her computer almost all day. She is left-handed. She does not smoke or use illicit drugs. She is in sexual relationship with two male partners and uses condoms.
**Clinical trial information**
**Description:** This study will look at the safety and tolerability of TPI 1100 in healthy volunteers and look at pharmacodynamic (PD) ···
**Inclusion Criteria:** - Healthy, male or female volunteers aged 18 to 55 years inclusive - Screening/baseline FEV1 greater than 90% predicted, - Body mass index (BMI) of 19 to 28 inclusive, - Clinical laboratory values and/or vital signs within normal reference ranges or not considered clinically significant by the Investigator
**Exclusion Criteria:** - Airways or systemic conditions that might affect respiratory function, including but not limited to clinically significant cardiac problems, - Breast-feeding or pregnancy, - Positive tests for smoking tobacco, alcohol, hepatitis B-surface antigen, hepatitis C antibody, and HIV at screening, - History of serious adverse reaction to any drugs.
**Q**: Please provide clinical trial matching result.
**A**: The patient is **irrelevant** for the clinical trial. ✗
**Reference**: The patient is **eligible** for the clinical trial.

---

**Patient information**
A 23-year-old woman comes to the emergency department with a history of nosebleeds lasting for 1 hour. She has a history of heavy menses as well as occasional gum bleeding following dental procedures. Her mother also has a history of menorrhagia. Laboratory tests reveal increased bleeding time and slightly increased partial thromboplastin time. She has no other medical conditions and is otherwise healthy. Her coagulation study shows CB = 0.30 IU/mL and FVIII:C = 0.37 IU/mL. She is not smoking or using any kind of illicit drugs. She uses alcohol occasionally and is in a relationship with her boyfriend for the past 2 years.
**Clinical trial information**
**Description:** The high percentage of failure using available non-surgical options to treat menorrhagia in women with bleeding disorders shows a continuing need for innovative treatments ···
**Inclusion Criteria:** - All menstruating women regardless of age - Women with heavy periods as measured by pictorial blood assessment chart - Women diagnosed with a bleeding disorder
**Exclusion Criteria:** - Acquired defective color vision - Factor VIII, Factor IX, FactorXI levels >250% - An inherited thrombophilic defect detected because of a positive family or personal history of thrombosis - Current use of oral contraceptives
**Q**: Please provide clinical trial matching result.
**A**: The patient is **eligible** for the clinical trial. ✗
**Reference**: The patient is **ineligible** for the clinical trial.

**Supplementary Fig. 9**: Examples of predictions by BiomedGPT in clinical-trial matching.

**Supplementary Table 1**: Comparison of performance (exact-match accuracy) between BiomedGPT and three other SOTA methods (MMQ [5], EKAID [6] and PLURAL [7]) on non-difference questions in MIMIC-Diff-VQA dataset.

| Model | Open Questions | Closed Questions | All Questions |
| --- | --- | --- | --- |
| BiomedGPT-B (ours) | 55.2 | 88.0 | 71.2 |
| MMQ | 11.5 | 10.8 | 11.5 |
| EKAID | 26.4 | 79.9 | 52.5 |
| PLURAL | 51.2 | 87.3 | 68.8 |

**Supplementary Table 2**: Fine-tuned experimental results of BiomedGPT on MedMNIST, a low-resolution (28×28) image classification dataset. We list the performance of the state-of-the-art (SOTA) approaches for each modality.

| Dataset | Domain / Modality | Metric | SOTA | | BiomedGPT | | |
| --- | --- | --- | --- | --- | --- | --- | --- |
| | | | Model | Result | Small | Medium | Base |
| MedMNIST | Colon pathology | Accuracy | MedViT-S (224) | 94.2 | 89.4 | 92.1 | 92.6 |
| | Dermatoscopy | | PMC-CLIP | 79.8 | 75.2 | 78.0 | 78.6 |
| | Retinal OCT | | MedViT-S (224) | 78.2 | 79.5 | 81.9 | 81.6 |
| | Chest X-Ray | | MedViT-S (224) | 96.1 | 91.8 | 93.4 | 96.7 |
| | Breast Ultrasound | | PMC-CLIP | 91.3 | 84.6 | 87.8 | 87.8 |
| | Blood Cell Microscope | | AutoML | 96.6 | 94.2 | 97.2 | 97.7 |
| | Axial Abdominal CT | | ResNet-18 (224) | 95.1 | 92.6 | 94.7 | 95.2 |
| | Coronal Abdominal CT | | MedViT-L (224) | 92.2 | 92.2 | 92.3 | 93.1 |
| | Sagittal Abdominal CT | | AutoKeras | 81.3 | 80.0 | 82.0 | 82.3 |

**Supplementary Table 3**: Different question categories of VQA in human evaluation.

| Question type | Example |
|---|---|
| Abnormality | What abnormalities are seen in the image? |
| Presence | Is there any evidence of <abnormality>? |
| Location | Where in the image is the <abnormality> located? |
| Type | What type is the <abnormality>? |
| Level | What level is the <abnormality>? |
| View | Which view is this image taken? |

**Supplementary Table 4**: Experimental details for the 3D extension of BiomedGPT: For MRI evaluation datasets—specifically AIBL [8] and MIRIAD [9]—we fine-tuned the model using data from the ADNI database (https://adni.loni.usc.edu). This dataset includes 1,216 images from normal controls (NC) and 1,565 images from Alzheimer's disease (AD) patients. For the LIDC [10] dataset, we fine-tuned the model on 1,050 training samples. For process 3D volumes, we integrated ACS Convolutions [11] into our framework. For image-only pre-training, we used our pre-trained 3D VQGAN based on BRaTS 2021 [12] and LUNA16 [13] datasets following [14].

| 3D image encoder | 3D VQGAN | Evaluation Dataset |
|---|---|---|
| ACS Convolutions-based | We pre-trained 3D VQGAN using BRaTS 2021 and LUNA16 datasets. Codebook: 16,384 visual token. Input dimension: $32 \times 256 \times 256$. Compression ratio: 8 (i.e., images of size $32 \times 256 \times 256$ have a latent dimension of $4 \times 32 \times 32$) | 1. AIBL: 363 NC cases and 50 AD cases. 2. MIRIAD: 177 NC and 346 AD cases. 3. LIDC: 66 benign and 33 malignant cases. |

**Supplementary Table 5**: Fine-tuning hyperparameters and settings for different tasks.

| Task / Data | | Image augmentation | Image resolution | Fine-tuning epochs | Maximum target length | Other hyper-parameters |
|---|---|---|---|---|---|---|
| Image classification | | Random resize cropping, random flipping, RandAug, random erasing. Furthermore, we incorporate Mixup and CutMix augmentations, each presenting a 50% chance of being applied to every batch. We set the alpha parameters for Mixup and CutMix at 0.8 and 1.0, respectively. | 384×384 for base, 256×256 for medium and small. | 5 | 30 | Dropout rate: 0.1<br><br>Label smoothing ratio: 0.1<br><br>Learning rate: 7e-5<br><br>Batch size: 128, 64, and 32 for small-, medium-, and base-size models, respectively. |
| Captioning | IU X-ray | Not implemented | 480×480 | 50 | 512 | By default, we set the beam search size to 3.<br><br>Learning rate: 7e-5 |
| | PEIR GROSS | | | | 20 | |
| | MIMIC-CXR | | | | 768 | |
| | ROCO | | | | 30 | |
| VQA | SLAKE | Not implemented | 384×384 for base, 256×256 for medium and small. | 50 | 128 | Learning rate: 7e-5<br><br>Inference type in validation:<br>1. SLAKE & PathVQA: beam search<br>2. VQA-RAD: all-candidate search |
| | PathVQA | | | 50 | | |
| | VQA-RAD | | | 100 | | |
| Text understanding | MedNLI | Not applicable | Not applicable | 20 | 30 | Learning rate: 7e-5<br><br>We continued training the model on the SEER, TREC'22, and MIMIC-III datasets, using the fine-tuned checkpoint on MedNLI as the starting point. |
| | SEER | | | 10 | 64 | |
| | TREC'22 | | | 10 | 12 | |
| | MIMIC-III | | | 10 | 30 | |
| Text summarization | MeQSum | Not applicable | Not applicable | 30 | 128 | Learning rate: 1e-4<br>Length penalty: 0.7<br>Label smoothing ratio: 0.1 |
| | HealthCareMagic | | | | | |
| | MIMIC-III | | | | | |
| | MIMIC-CXR | | | | | |

**Supplementary Table 6**: Preliminary ablation study of assessing the influence of two hyper-parameters for image captioning task (on IU X-RAY dataset). Note that the maximum target length is too small, and not appropriate.

| Maximum target length | Beam search size | CIDEr |
|---|---|---|
| 20 | 5 | 16.4 |
| 20 | 10 | 29.4 |
| 20 | 20 | 29.8 |
| 40 | 10 | 35.1 |
| 45 | 10 | 36.0 |
| 50 | 10 | 31.1 |
| 60 | 10 | 31.5 |

**Supplementary Table 7**: Train/Validation/Test set splits and referenced pre-processing sources for fine-tuning datasets used in BiomedGPT.

| Task | Dataset | Train / valid / test split | Metric for checkpoint selection | Reference |
|---|---|---|---|---|
| Image classification | MC-CXR | 96 / 13 / 29 | Accuracy | [15] |
| | SZ-CXR | 463 / 66 / 133 | Accuracy | [15] |
| | CBIS-DDSM-CALC | 1,544 / 0 / 326 | Accuracy | [16] |
| | CBIS-DDSM-MASS | 1,526 / 0 / 378 | Accuracy | [16] |
| | MMR-Colon Pathology | 89,996 / 10,004 / 7,180 | Accuracy | [17] |
| | MMR-Dermatoscopy | 7,007 / 1,003 / 2,005 | Accuracy | [17] |
| | MMR-Retinal OCT | 1,080 / 120 / 400 | Accuracy | [17] |
| | MMR-Chest X-ray | 4,708 / 524 / 624 | Accuracy | [17] |
| | MMR-Breast Ultrasound | 546 / 78 / 156 | Accuracy | [17] |
| | MMR-Blood Cell Microscope | 11,959 / 1,712 / 3,421 | Accuracy | [17] |
| | MMR-Coronal Abdominal CT | 13,000 / 2,392 / 8,268 | Accuracy | [17] |
| Captioning | MIMIC-CXR | 117,671 / 919 / 1,530 | CIDEr | [18] |
| | PEIR GROSS | 5,359 / 1,339 / 745 | CIDEr | [19] |
| | IU X-RAY | 4,138 / 1,034 / 1,289 | CIDEr | [19] |
| VQA | SLAKE | 4,919 / 1,053 /1,061 | Accuracy | [20] |
| | PathVQA | 19,755 / 6,279 / 6,761 | Accuracy | [21] |
| | VQA-RAD | 1,797 / 0 / 451 | Accuracy | [22] |
| Text understanding | MedNLI | 11,232 / 1,395 / 1,422 | Accuracy | [23] |
| | SEER | 18724 / 0 / 4680 (10-Fold) | Accuracy | [24] |
| | TREC'22 | 28315 / 0 / 7079 (10-Fold) | Accuracy | N/A |
| | MIMIC-III | 509,311 / 73,621 / 147,331 | Accuracy | [25] |
| Text summarization | MeQSum | 800 / 100 / 100 | ROUGE-L | [26] |
| | HealthCareMagic | 181,119 / 22,641 / 22,642 | ROUGE-L | [26] |
| | MIMIC-CXR | 125,417 / 991 / 1,624 | ROUGE-L | [27] |
| | MIMIC-III | 59,320 / 7,413 / 6,526 | ROUGE-L | [27] |

# Supplementary reference